\documentclass[pmlr,twocolumn,10pt]{jmlr} 

\mlhtrack{proceedings}

\newif\iffinal
\finaltrue  

\iffinal
    \ifmlhneedspmlr
      \jmlrvolume{346}
      \jmlryear{2026}
    \fi
    \ifmlhfindings \jmlrproceedings{}{ML4H 2026 - Findings Track}\fi
    \ifmlhdemo     \jmlrproceedings{}{ML4H 2026 - Demo Track}\fi
    \jmlrworkshop{Machine Learning for Health (ML4H) 2026}
\else
    \jmlrproceedings{}{Submitted to ML4H 2026: \mlhtrackname}
    \jmlrworkshop{Machine Learning for Health (ML4H) 2026}
\fi

\usepackage{booktabs}
\usepackage{siunitx}

\usepackage[switch]{lineno}

\theorembodyfont{\upshape}
\theoremheaderfont{\scshape}
\theorempostheader{:}
\theoremsep{\newline}

\usepackage{multirow}
\usepackage{multicol}
\usepackage{rotating}
\usepackage{listings}
\usepackage{fancyvrb}
\usepackage[cachedir=_minted, frozencache]{minted}
\setminted{encoding=utf8}
\usepackage{nicefrac}
\usepackage{microtype}
\usepackage{float}
\usepackage{longtable}
\usepackage{lscape}
\usepackage{threeparttablex}
\usepackage{dsfont}

\newcommand {\hi}[1] {\textcolor{blue}{\bf #1}}

\title[Guideline-Based Medical Reasoning]{Guideline-Based Medical Reasoning in LLMs}

\author{%
 \Name{Michael Staniek} \Email{staniek@cl.uni-heidelberg.de}\\
  \addr Department of Computational Linguistics, Heidelberg University, Germany
 \AND
 \Name{Artem Sokolov\nametag{\thanks{Participated in advisory capacity.}}} \Email{sokolov@cl.uni-heidelberg.de}\\
 \addr Google DeepMind, Berlin, Germany
 \AND
 \Name{Stefan Riezler} \Email{riezler@cl.uni-heidelberg.de}\\
 \addr Department of Computational Linguistics, Heidelberg University, Germany \\\ \& Interdisciplinary Center for Scientific Computing (IWR), Heidelberg University, Germany
}

\begin{document}

\maketitle

\ifmlhdemo\else
\begin{abstract}
Large Language Models (LLMs) excel at medical question-answering, yet progress toward medical reasoning models is limited by the lack of scalable, trustworthy reasoning supervision. We propose tapping into the so far underused resource of medical consensus guidelines. These map numerical clinical measurements to disease scores and can optionally be curated for site-specific exceptions.  Automatically generating gold-standard reasoning chains from consensus rules enables both large-scale supervised fine-tuning and automatic evaluation. Applying SOFA, SAPS-II, and Sepsis-3 guidelines to MIMIC-III, we show that small fine-tuned models significantly outperform larger one-shot prompted models in derivation correctness and F1 score, achieving near-perfect deductive inference and reliable exception handling. Trustworthy step-by-step reasoning does not reliably emerge from one-shot prompting, even when models receive the consensus definition and an example. Performance across all models declines when forecasting future clinical variables, indicating that, on the evaluated structured ICU tasks, the central challenge is not rule application, but the orthogonal problem of generalizing into the future for prognostic reasoning.

\end{abstract}
\begin{keywords}
medical reasoning, medical consensus guidelines, reasoning LLMs
\end{keywords}
\fi

\ifmlhneedsstatements
\paragraph*{Data and Code Availability}
This work uses public data from physionet.org. Code can be found here: \url{https://github.com/StatNLP/guideline_based_medical_reasoning_LLM}.

\paragraph*{Institutional Review Board (IRB)}
IRB approval is not required for public data.
\fi

\section{Introduction}
\label{sec:intro}

Current research on medical reasoning with LLMs largely relies on fine-tuning pretrained models on medical question-answering data \citep{XieETAL:25,SinghalETAL:25}, with reasoning processes expected to emerge through chain-of-thought prompting \citep{WeiETAL:23}. This paradigm has been extended by fine-tuning on manually crafted medical reasoning chains \citep{LiuETALgeneralist:25} and by reinforcing reasoning behavior using rewards based on final-answer correctness \citep{ChenETALhuatuo:24,ZhangETALmedrlvr:25,LiuETALdistillation:25}. Consequently, medical reasoning is typically evaluated through final-answer accuracy on question-answering benchmarks or through assessment by human experts (see \cite{WangETALsurvey:25} for a recent survey).

Since manually crafting medical reasoning chains does not scale, large-scale fine-tuning and automatic evaluation of the inference processes themselves remain difficult. Our work argues that medical consensus guidelines can fill this gap. We propose automatically verbalizing guideline instantiations on electronic health records and, where needed, curating them to capture patient-cohort-specific exceptions. Teaching LLMs to follow medical inference\footnote{To avoid the anthropomorphic attribution of "reasoning" capabilities to LLMs, we use the terms medical "rules" and "inference processes" throughout.} rules and their exceptions step-by-step
goes beyond the deterministic application of guideline rules and requires associative learning. Fine-tuning LLMs on curated, verbalized inference chains provides such supervision while enabling automatic evaluation against gold-standard medical inference chains. Crucially, the goal is not to replace a calculator. 
While rule engines and logic programs \citep{Lifschitz:08} apply fixed rules exactly, or inductive logic programming \citep{LawETAL:20}  learns exceptions from corrected examples, we study a different scenario where a physician corrects LLM outputs with site-specific exceptions.  Since there is no deterministic consensus that specifies which and how many preconditions alter scores like SOFA, the LLM needs to learn such corrections \emph{associatively} from examples while producing a verbalized inference chain. Our experiments simulate such corrections since no clinician corrections of LLM outputs are available.

Medical consensus guidelines are evidence-based recommendations developed by representative groups of experts for the diagnosis and treatment of disease. They exist across nearly all areas of medicine, ranging from mental disorders \citep{DSMV:13} and neurological diseases \citep{McDonaldETAL:01} to physiological conditions such as kidney injury \citep{KDIGO:12}, organ failure \citep{VincentETAL:96,LeGallETAL:93}, and sepsis \citep{SeymourETAL:16,SingerETAL:16}. In this work, we focus on guidelines with deterministic rules such that the related gold-standard inference chains for training and evaluation are automatically verifiable. Guidelines that rest on clinical judgment, imaging, or symptom descriptions are outside the scope of this paper.
In our experiments, we focus on three complex compositional consensus definitions: the Sequential Organ Failure Assessment (SOFA), Sepsis-3, and the Simplified Acute Physiology Score (SAPS-II). We selected these definitions because they share basic inference rules in the form of \textit{step functions} that map clinical measurements to severity scores, while introducing complementary forms of complexity. SOFA incorporates \textit{exceptions to the rule}, where organ-specific scores are modified based on interventional therapies or ICD codes indicating relevant comorbidities. Sepsis-3 additionally requires computing a change in SOFA score of at least two points, which in an early-prediction setting requires time-series forecasting (TSF) of clinical variables 24 hours into the future. It therefore extends deterministic \textit{deductive inference} with \textit{inductive inference}, where the relation between premise and conclusion is probabilistic. Finally, SAPS-II introduces compositional complexity by combining step-function scores through a \textit{non-linear function}.

We investigate whether LLMs can be taught to perform step-by-step inference according to such complex medical consensus definitions. To this end, we automatically instantiate SOFA, Sepsis-3, and SAPS-II on public-domain patient time-series data and additionally annotate selected exceptions to their rules. We use the resulting inference chains to fine-tune and evaluate several open-source LLMs.\footnote{We use open-source models with up to 70B parameters. Comparison with larger closed-source commercial LLMs is prohibited by the PhysioNet data privacy policy.}

Our results show that small fine-tuned models (LLaMA 8B) consistently outperform substantially larger models using one-shot prompting with the explicit consensus definition (LLaMA 70B), models pretrained on medical text including the original consensus definitions (Me-LLaMA 8B), and specialized reasoning LLMs (Qwen 27B) in derivation correctness and in our main metric, the Sepsis-3 F1 score, although not on every individual metric (Table~\ref{tab:results_value_sofa}). In particular, fine-tuned LLMs achieve near-perfect generalization to unseen patients in the correct application of rules and exceptions, including synthetic exceptions under \emph{covariate shift}, in which held-out ICD codes from known ICD classes occur only at test time (Table~\ref{tab:synt_icd_results}). This associative capability, which goes beyond arithmetic rule application, does not emerge reliably in one-shot LLMs, even when the consensus definition and an example are provided in the prompt.

Performance nevertheless degrades across all models when future clinical variables must be forecast for early prediction. On the evaluated tasks, the ease of automatically generating in-distribution fine-tuning data, together with the near-perfect results for deductive inference, suggests that the primary bottleneck is not out-of-distribution rule application, but the orthogonal challenge of \emph{generalization into the future}: the inherent difficulty of time-series forecasting in prognostic applications of LLMs. This limitation may be addressed by moving beyond purely text-based representations toward multimodal architectures that incorporate dedicated time-series forecasting models.

\section{Related Work}
\label{sec:related}

Most training and evaluation pipelines for medical reasoning are based on medical question-answering \citep{LiuETAL:23,SinghalETAL:25}, while it has been shown that the vast majority of questions in medical QA benchmarks can be answered by factual recall, not requiring multi-step inference at all \citep{ThapaETAL:25}. 
While excelling in medical question-answering tasks, few-shot LLMs suffer from prompt brittleness \citep{HochlehnertETAL:25,ReeseETAL:24,GohETAL:24,BeanETAL:26}, fail in basic TSF tasks \citep{MerrillETAL:24,TanETAL:24}, and lack adherence to consensus guidelines \citep{HagerETAL:24}. 
Various approaches have been presented to teach LLMs medical inference rules, e.g., by pretraining or finetuning on medical texts including consensus guidelines \citep{ChenETALmeditron:23,XieETAL:25}, by learning from manually crafted demonstrations \citep{LiuETALgeneralist:25} or from knowledge graphs \citep{WuETALmedreason:25},  or by eliciting emergent reasoning via reinforcement learning \citep{ChenETALhuatuo:24,ZhangETALmedrlvr:25,LiuETALdistillation:25}. However, with the exception of \cite{FanETALchestx:25} and \cite{ZhouETAL:25} who use automatic evaluation metrics or LLM-as-Judge approaches to compare predicted against expert-crafted reasoning chains, the state-of-the-art in evaluation of medical reasoning is with respect to final-answer accuracy on standard medical question-answering benchmarks. This includes work like MedGUIDE where cancer-related guidelines are converted to multiple-choice diagnostic questions \citep{LiETALmedguide:25}. Our work relies on automatic generation and curation of gold standard medical inference chains, which minimizes manual labor and yet allows supervised training and automatic evaluation of full medical inference processes.

\begin{figure*}[t!]
    \centering
    \includegraphics[width=0.46\textwidth]{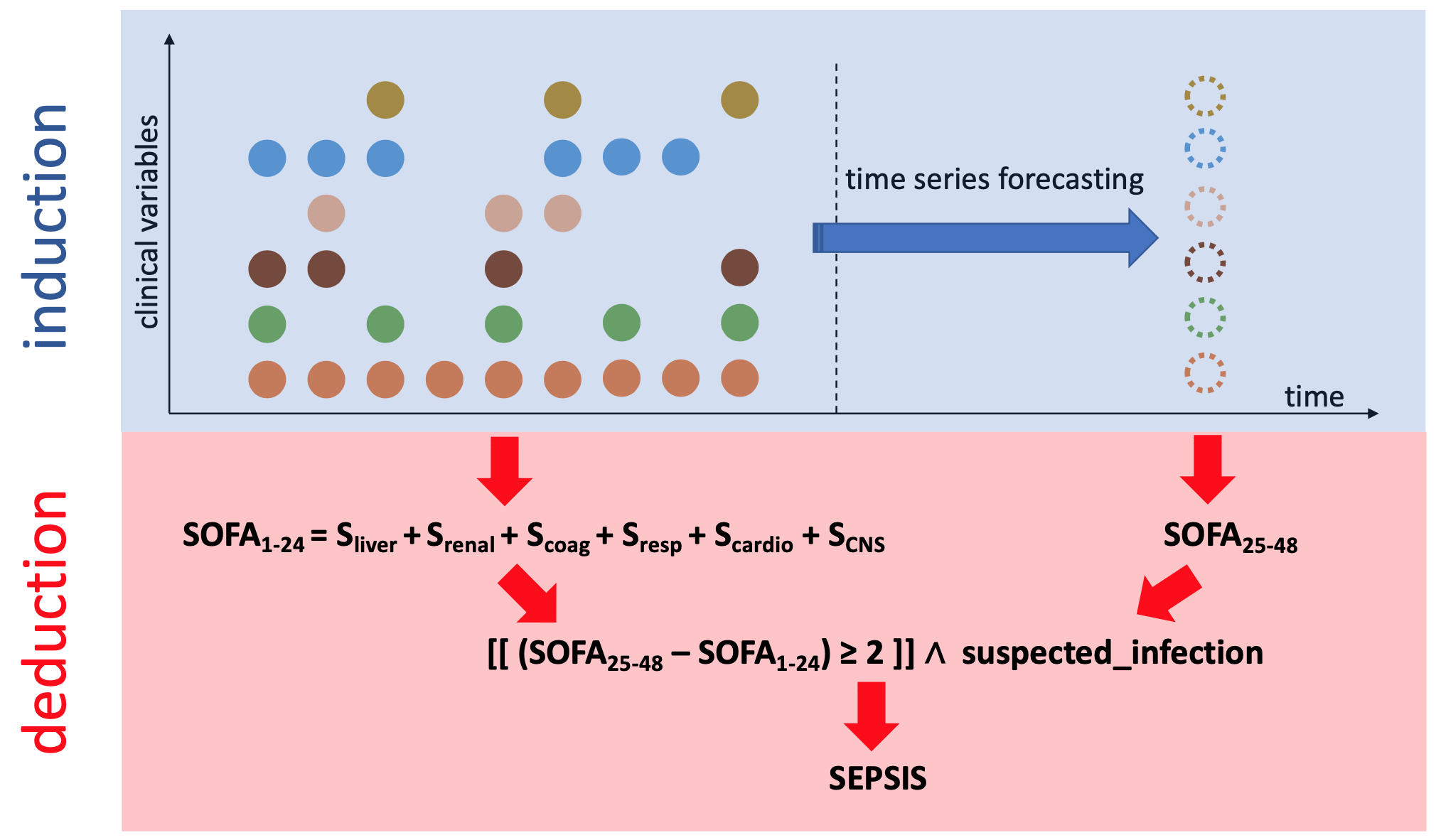}
    \caption{Inductive and deductive inference rules in the Sepsis-3 Definition \citep{SingerETAL:16,SeymourETAL:16}. Deductive rules calculate extrema over time, map thresholds onto step functions for SOFA scores, and calculate total SOFA and changes over time. Inductive rules involve TSF of clinical variables 24 hours into the future.} 
    \label{fig:compositional}
\end{figure*}

\section{Teaching Medical Inference Rules to LLMs}
\label{sec:inference}

Medical inference covers the following non-trivial steps: \emph{Deductive inference} covers the extraction of worst values, mapping them to step/logit functions, and applying \emph{exceptions} that tie premises like ICD codes to subscore changes, while \emph{inductive inference} is the probabilistic step of forecasting future values before those rules apply; it is the exceptions and the forecasting, not the arithmetic, that make the task non-trivial.

\paragraph{Deductive Inference}

SOFA \citep{VincentETAL:96}, Sepsis-3 \citep{SingerETAL:16,SeymourETAL:16}, and SAPS-II \citep{LeGallETAL:93}, each are based on two deductive inference steps: A first step is a calculation of "worst" values (minima or maxima) of clinical variables over 24 hours. A second step is a deductive syllogism consisting of rules (the major premises) that are instantiated to thresholds on clinical variables (the minor premises) that are mapped to step functions (the conclusions). For SAPS-II, these consist of 12 non-static physiological variables (see Appendix \ref{appendix:saps}) each of which is mapped to a score between $0$ and $26$, which are summed up to a total score, in addition to scores for age, admission type, and three chronic disease variables. SOFA consists of six subscores from $0-4$, each corresponding to an organ system, namely the central nervous system, and the cardiovascular, respiratory, coagulation, liver, and renal organ systems (see Appendix \ref{appendix:sofa}). The total SOFA score is calculated by summing these six subscores to a score ranging from $0-24$.
The deductive rules of Sepsis-3 (see the lower part of Figure \ref{fig:compositional}) calculate SOFA once for clinical measurements during the first 24 hours (SOFA$_{1-24}$), and for predicted measurements in the next 24 hours (SOFA$_{25-48}$). A further deductive step computes the change indicator $\text{SOFA}_\text{diff} := [\![ (\text{SOFA}_{25:48} - \text{SOFA}_{1:24}) \geq 2 ]\!]$, where $[\![ a ]\!]=1$ if $a$ is true and $0$ otherwise, and the binary Sepsis label combines it with a suspected-infection indicator\footnote{Following \cite{SingerETAL:16,SeymourETAL:16}, a suspected infection is defined as a combination of antibiotics treatment and blood cultures, starting within the first 24 hours after admission.}: $\text{SEPSIS} := \text{SOFA}_\text{diff} \wedge \text{suspected$\_$infection}$.

\paragraph{Inductive Inference}

Inductive inference is a necessary component of any prognostic application of medical guidelines.
Sepsis-3 is laid out to detect a life-threatening organ dysfunction by an acute \textit{change} in the total SOFA score, thus an inductive component involving TSF of future values of clinical features is required (see upper part of Figure \ref{fig:compositional}). This forecast is intrinsic to Sepsis-3, not an artifact of our setup: flagging sepsis at time $t$ needs SOFA at both $t$ and $t{+}x$, and the latter requires forecasting for \emph{early} (non-retrospective) prediction. For the case of Sepsis-3, 131 clinical measurements extracted from the MIMIC-III database \citep{JohnsonETAL:16} are used (see Appendix \ref{appendix:features}). 
The TSF task in our experiments is defined to predict the next 24 hours from a history of preceding 24 hours. Dedicated models to perform the TSF task are described in Appendix \ref{appendix:tsf}.

\paragraph{Handling Exceptions to the Rules}

We implement a scenario where verbalized rule instantiations for a particular patient cohort are manually curated for exceptions that arise because of chronic comorbidities or interventional therapies. Teaching exceptions to the rules goes beyond a deterministic application of guideline rules and constitutes an associative learning task where special medical conditions have to be mapped to deterministic increases in disease severity scores. We present experiments where we use a selected list of exceptions to SOFA guidelines from the literature \citep{CentnerETAL:20,AlrawashdehETAL:24,RanzaniETAL:25} and modify the medical inference chains for patients with the respective ICD codes or therapies. The full table of selected exceptions, and a histogram of occurrences of these exceptions in the training data, is given in Table \ref{tab:icd9_mappings} and Figure \ref{fig:icd9_histogram} in Appendix \ref{appendix:exc}. These literature-derived exceptions are evaluated in-distribution. Furthermore, in order to control for out-of-distribution generalization, we present a separate experiment with \emph{synthetic} exceptions, where ICD codes are randomly assigned to data, but some codes from the same ICD class are only seen at test time. The distribution shift addressed here is \emph{covariate shift}: the marginal distribution of the input codes differs between training and test data, while the mapping from ICD class to exception stays the same. This tests generalization within a constructed exception schema, not generalization to clinically novel exceptions.
A definition and statistics for synthetic exceptions are given in Tables \ref{tab:synth_icd} and \ref{tab:synth_icd_histogram} in Appendix \ref{appendix:exc}.

\paragraph{Mapping to Non-Linear Functions}

SAPS-II departs from the step-function regime of the other guidelines: on top of the thresholded scores, a logit function of the total score is computed and mapped through a sigmoid to a probability of hospital mortality. The exact logit and sigmoid, with the fitted coefficients of the SAPS-II definition \citep{LeGallETAL:93}, are given in Appendix~\ref{appendix:saps}.

\paragraph{Teaching Medical Inference by Finetuning} All inference rules are taught to LLMs via supervised finetuning, where an input time series, a verbalized instantiation of the inference rules (and exceptions) to patient data, and the answer, is provided to the model during training. In our case, the verbalized instantiation for SOFA and Sepsis-3 is done by a manually written template (included in the code-base). The verbalization code for SAPS-II was written by Claude\footnote{\url{http://claude.ai}} given our hand-written Sepsis-3 verbalization code and the Sepsis-3 and SAPS-II consensus definitions. Examples for finetuning data following SAPS-II and Sepsis-3, respectively, are given in Appendix \ref{appendix:verbalization} and \ref{appendix:finetune-prompt-sapsii}. 

\section{Evaluating Medical Inference in LLMs}
\label{sec:metrics}

All three evaluations below are computed \emph{fully automatically}, with no human annotation of outputs.


\paragraph{Derivation Correctness} First, we evaluate the LLM output according to derivation correctness  which checks whether each child node (conclusion) in the predicted inference graph follows from the parent node (premise) according to the consensus rule.
This evaluation measures the ability of the LLM to learn the deductive inference rules of the consensus definition. Furthermore, it checks the internal consistency of the verbalized inference chain. It does not prove that the chain reflects the computation behind the model's prediction, and it partly depends on adherence to the output format, since outputs are parsed automatically. Derivation correctness for SOFA and Sepsis-3 is illustrated in Appendix Figure \ref{fig:evaluation-metrics} (dashed blue arrows). A calculation for SAPS-II is done analogously.

\paragraph{Forced Derivation Correctness} In order to avoid trivial cases of derivation correctness, for example, inference of unchanged SOFA scores from unchanged future measurement predictions, we measure in addition forced derivation correctness. This is done by forcing the decoder to use the correct inference chain up to a last measurement token of a SOFA system as context, then continuing the decoding process until the specified target node.

\paragraph{Value Correctness} Furthermore, we conduct an evaluation according to value correctness which compares the numerical value of each partial inference node in the prediction graph with a corresponding real-world output. 
Since errors earlier in the inference necessarily propagate through the inference chain, this evaluation measures the utility of the LLM output in practical applications. An illustration for SOFA and Sepsis-3 is given in Appendix Figure \ref{fig:evaluation-metrics} (dotted red arrows). A calculation for SAPS-II is done analogously.

\paragraph{Numerical Metrics} All evaluations are computed automatically by keyword parsing and number extraction with a $5\%$ tolerance; details for the tolerance are given in Appendix~\ref{appendix:margin}. For value correctness, we compare predicted Sepsis-3 labels against labels derived from real-world measurements in the second 24-hour block and compute accuracy, specificity, sensitivity, and F1. All finetuning improvements were confirmed for statistical significance over one-shot learning with an approximate randomization test (Appendix~\ref{appendix:evaluation-metrics}).

\section{Experimental Setup}
\label{sec:exps}
       
The basic LLM used in our experiments is a pretrained Llama-3 model with 8B parameters \citep{GrattafioriETAL:24}\footnote{\scriptsize\url{https://huggingface.co/meta-llama/Llama-3.1-8B-Instruct}} used in \textbf{one-shot} mode.
All one-shot experiments use the prompts shown in Appendices \ref{appendix:one-shot}, \ref{appendix:one-shot-exception}, and \ref{appendix:one-shot-sapsii}, that include the explicit SAPS-II, SOFA, and Sepsis-3 rules together with an example instantiation to patient data, respectively.
Our \textbf{finetuned} model uses LORA adapters \citep{HuETAL:22lora} on verbalizations of inference processes. Examples for finetuning data are given in Appendices \ref{appendix:verbalization} and \ref{appendix:finetune-prompt-sapsii}, showing verbalized instantiations of SAPS-II, SOFA, and Sepsis-3 rules to patient data, respectively, without explicit verbalizations of the abstract inference rules or exceptions.
Furthermore, we use a dedicated medical TSF forecaster, pretrained on MIMIC-III to improve the inductive inference part of the LLM (see Appendix \ref{appendix:tsf}).
The forecaster follows \citet{StaniekETAL:24mlhc}, with the best hyperparameters from that paper (code: \url{https://github.com/StatNLP/mlhc\_2024\_prediction\_of\_causes/}).
The TSF model is used in two ways: in the \textbf{pipeline} approach we augment the prompt with the forecaster's predictions while finetuning Llama-3 with LORA and keeping the forecaster fixed; in the \textbf{multimodal} approach we couple the two, prepending the forecaster's decoder outputs (mapped by a connector MLP to the Llama-3 embedding dimension) to the text embeddings and updating the LORA adapters, the connector MLP, and the TSF model jointly (architecture detail in Appendix~\ref{appendix:tsf}).
A list of hyperparameter settings chosen on the validation set (except the default setting for the Llama-3 8B model) is given in Appendix \ref{appendix:hyperparameters}. 
For further comparison, we use an 8B parameters LLM  pretrained on medical texts, including publications and wikipedia texts on consensus guidelines (\textbf{me-llama} \citep{XieETAL:25}\footnote{ \scriptsize\url{https://huggingface.co/YBXL/Med-LLaMA3-8B}}), two reasoning-style LLMs (\textbf{deepseek} \citep{DeepSeek:25}\footnote{ \scriptsize\url{https://huggingface.co/deepseek-ai/DeepSeek-R1-Distill-Llama-8B}} and \textbf{qwen3.6-27B} \citep{qwen3.6-27b}\footnote{ \scriptsize\url{https://huggingface.co/Qwen/Qwen3.6-27B}} ), and a 70B pretrained Llama-3 model (\textbf{one-shot-70B}\footnote{ \scriptsize\url{https://huggingface.co/meta-llama/Llama-3.1-70B-Instruct}}), all in one-shot mode (few-shot would have exceeded context length restrictions). We additionally evaluated HuatuoGPT-o1-8B and Olmo-3-7B-Instruct under the identical one-shot prompts (Appendix~\ref{appendix:further-llms}).

The data used in our experiments are electronic health records from MIMIC-III ( see Appendix \ref{appendix:data} for details).

\begin{table*}[t!]
    \centering
\resizebox{0.95\textwidth}{!}{
    \begin{tabular}{lllllll|lll}
    \toprule
 &  variable & one-shot & one-shot-70B & me-llama & deepseek & qwen3.6-27B & finetuned & pipeline & multimodal \\
\midrule
\multirow{16}{*}{\begin{sideways}SAPS-II current\end{sideways}}
& S$_\text{Age} $        & 0.138 & 0.200 & 0.133 & 0.094 & 0.171 & 1.000 & & \\
& S$_\text{HR} $          & 0.322 & 0.548 & 0.629 & 0.551 & 0.845 & 1.000 & & \\
& S$_\text{SBP} $         & 0.270 & 0.487 & 0.298 & 0.059 & 0.862 & 1.000 & & \\
& S$_\text{Temperature} $ & 0.797 & 0.761 & 0.942 & 0.750 & 0.930 & 0.997 & & \\
& S$_\text{GCS}$          & 0.533 & 0.795 & 0.955 & 0.477 & 0.878 & 1.000 & & \\
& S$_\text{PaO2/FiO2}$    & 0.434 & 0.286 & 0.622 & 0.326 & 0.287 & 0.999 & & \\
& S$_\text{BUN}$          & 0.595 & 0.752 & 0.682 & 0.550 & 0.825 & 1.000 & & \\
& S$_\text{Urine} $       & 0.452 & 0.565 & 0.279 & 0.434 & 0.889 & 1.000 & & \\
& S$_\text{Sodium} $      & 0.824 & 0.571 & 0.872 & 0.537 & 0.727 & 1.000 & & \\
& S$_\text{Potassium} $   & 0.862 & 0.813 & 0.888 & 0.647 & 0.846 & 1.000 & & \\
& S$_\text{Bicarbonate}$  & 0.851 & 0.790 & 0.876 & 0.667 & 0.762 & 1.000 & & \\
& S$_\text{WBC} $         & 0.866 & 0.763 & 0.891 & 0.686 & 0.839 & 1.000 & & \\
& S$_\text{Bilirubin}$    & 0.903 & 0.429 & 0.925 & 0.725 & 0.502 & 1.000 & & \\
& Total SAPS-II     & 0.532 & 0.322 & 0.597 & 0.013 & 0.222 & 1.000 & & \\
& Logit             & 0.065 & 0.083 & 0.166 & 0.024 & 0.074 & 1.000 & & \\
& Probability       & 0.301 & 0.859 & 0.750 & 0.554 & 0.947 & 1.000 & & \\
\midrule
\multirow{7}{*}{\begin{sideways}SOFA current\end{sideways}} 
& S$_\text{CNS}$       & 0.693 & 0.803 & 0.520 & 0.288 & 0.983 & 1.000 &  &  \\
& S$_\text{cardio}$    & 0.516 & 0.696 & 0.258 & 0.338 & 0.822 & 0.998 &  &  \\
& S$_\text{resp}$      & 0.561 & 0.302 & 0.404 & 0.279 & 0.449 & 0.995 &  &  \\
& S$_\text{coag}$      & 0.599 & 0.763 & 0.511 & 0.381 & 0.857 & 0.966 &  &  \\
& S$_\text{liver}$     & 0.537 & 0.864 & 0.601 & 0.336 & 0.899 & 0.993 &  &  \\
& S$_\text{renal}$     & 0.537 & 0.730 & 0.396 & 0.256 & 0.829 & 1.000 &  &  \\
& $\text{SOFA}_{1:24}$ & 0.543 & 0.893 & 0.777 & 0.165 & 0.800 & 1.000 &  &  \\
\midrule
\multirow{8}{*}{\begin{sideways}SOFA future\end{sideways}} 
& S$_\text{CNS}$            & 0.504 (0.827) & 0.545 (0.895) & 0.399 (0.786) & 0.261 (0.823) & 0.836 (0.989) & 1.000 (1.000) & 1.000 (1.000) & 1.000 (1.000) \\
& S$_\text{cardio}$         & 0.580 (0.694) & 0.679 (0.714) & 0.353 (0.647) & 0.269 (0.671) & 0.818 (0.846) & 0.987 (0.997) & 0.975 (0.997) & 0.965 (0.997) \\
& S$_\text{resp}$           & 0.530 (0.756) & 0.307 (0.827) & 0.384 (0.733) & 0.266 (0.722) & 0.445 (0.891) & 0.995 (0.996) & 0.996 (0.996) & 0.995 (0.996) \\
& S$_\text{coag}$           & 0.661 (0.786) & 0.754 (0.818) & 0.519 (0.801) & 0.306 (0.747) & 0.822 (0.896) & 0.967 (1.000) & 0.970 (0.999) & 0.966 (1.000) \\
& S$_\text{liver}$          & 0.534 (0.835) & 0.847 (0.969) & 0.599 (0.657) & 0.335 (0.776) & 0.896 (0.833) & 0.992 (0.999) & 0.992 (0.999) & 0.992 (1.000) \\
& S$_\text{renal}$          & 0.554 (0.708) & 0.737 (0.773) & 0.414 (0.690) & 0.251 (0.634) & 0.800 (0.773) & 1.000 (1.000) & 1.000 (1.000) & 1.000 (1.000) \\
& $\text{SOFA}_{25:48}$     & 0.518 (0.875) & 0.868 (0.984) & 0.768 (0.992) & 0.148 (0.550) & 0.849 (0.998) & 1.000 (1.000) & 0.999 (1.000) & 1.000 (1.000) \\
& $\text{SOFA}_\text{diff}$ & 0.845 (0.740) & 0.891 (0.477) & 0.769 (0.414) & 0.503 (0.351) & 0.894 (0.746) & 1.000 (1.000) & 1.000 (1.000) & 1.000 (1.000) \\
\midrule
\multirow{1}{*}{\begin{sideways}\end{sideways}}
& SEPSIS & 0.859 (0.682) & 0.909 (0.727) & 0.769 (0.688) & 0.738 (0.521) & 0.972 (0.684) & 1.000 (1.000) & 1.000 (1.000) & 1.000 (1.000) \\
\bottomrule
    \end{tabular}
}
\caption{Derivation correctness of predictions. Forced derivation correctness in brackets. Pipeline and multimodal columns apply only to the SOFA-future and SEPSIS blocks; SAPS-II and SOFA-current use the current 24-hour window (no forecasting), so those variants coincide with the finetuned model and the cells are intentionally left blank.}
    \label{tab:results_inference}
\end{table*}
\begin{table*}[t!]
    \centering
\resizebox{0.7\textwidth}{!}{
    \begin{tabular}{lllllll|l}
    \toprule
 &  variable & one-shot & one-shot-70B & me-llama & deepseek & qwen3.6-27B & finetuned \\
\midrule
\multirow{7}{*}{\begin{sideways}SOFA current\end{sideways}} 
& S$_\text{CNS}$       & 0.810 & 0.788 & 0.652 & 0.450 & 0.987 & 0.989 \\
& S$_\text{cardio}$    & 0.416 & 0.627 & 0.301 & 0.361 & 0.767 & 0.999 \\
& S$_\text{resp}$      & 0.577 & 0.342 & 0.542 & 0.323 & 0.636 & 0.994 \\
& S$_\text{coag}$      & 0.562 & 0.649 & 0.555 & 0.244 & 0.784 & 0.945 \\
& S$_\text{liver}$     & 0.626 & 0.855 & 0.847 & 0.555 & 0.942 & 0.996 \\
& S$_\text{renal}$     & 0.153 & 0.202 & 0.133 & 0.384 & 0.444 & 0.998 \\
& $\text{SOFA}_{1:24}$ & 0.607 & 0.907 & 0.965 & 0.257 & 0.891 & 0.999 \\
\midrule
\multirow{9}{*}{\begin{sideways}SOFA future\end{sideways}} 
& S$_\text{CNS}$            & 0.813 (0.839) & 0.791 (0.869) & 0.693 (0.780) & 0.457 (0.830) & 0.980 (0.980) & 0.988 (1.000) \\
& S$_\text{cardio}$         & 0.547 (0.739) & 0.637 (0.711) & 0.347 (0.670) & 0.349 (0.692) & 0.764 (0.842) & 0.984 (0.997) \\
& S$_\text{resp}$           & 0.583 (0.775) & 0.358 (0.781) & 0.517 (0.737) & 0.313 (0.729) & 0.636 (0.892) & 0.996 (0.996) \\
& S$_\text{coag}$           & 0.463 (0.786) & 0.641 (0.852) & 0.522 (0.777) & 0.164 (0.698) & 0.758 (0.879) & 0.948 (1.000) \\
& S$_\text{liver}$          & 0.659 (0.773) & 0.857 (0.959) & 0.849 (0.550) & 0.555 (0.530) & 0.922 (0.716) & 0.995 (0.999) \\
& S$_\text{renal}$          & 0.327 (0.760) & 0.151 (0.764) & 0.135 (0.701) & 0.383 (0.725) & 0.408 (0.758) & 0.999 (1.000) \\
& $\text{SOFA}_{25:48}$     & 0.637 (0.904) & 0.905 (0.985) & 0.959 (0.986) & 0.274 (0.569) & 0.917 (0.999) & 1.000 (1.000) \\
& $\text{SOFA}_\text{diff}$ & 0.596 (0.725) & 0.744 (0.584) & 0.592 (0.676) & 0.857 (0.520) & 0.854 (0.647) & 1.000 (1.000) \\
\midrule
 & SEPSIS & 0.736 (0.619) & 0.985 (0.682) & 0.785 (0.380) & 0.944 (0.359) & 0.931 (0.705) & 1.000 (1.000) \\
\bottomrule
    \end{tabular}
}
\caption{Derivation correctness of predicted values with data-driven exceptions to SOFA rules. Forced derivation correctness in brackets.}
    \label{tab:results_inference_exception}
\end{table*}

\begin{table*}[t!]
    \centering
    \resizebox{0.7\textwidth}{!}{
    \begin{tabular}{lllllll|lll}
    \toprule
 & variable & one-shot & one-shot-70B & me-llama & deepseek & qwen3.6-27B & finetuned & pipeline & multimodal \\
\midrule
\multirow{13}{*}{\begin{sideways}SAPS-II current\end{sideways}} 
& Age         & 1.000 & 0.972 & 1.000 & 0.903 & 1.000 & 1.000 & & \\
& HR          & 0.618 & 0.600 & 0.342 & 0.188 & 0.435 & 0.998 & & \\
& SBP         & 0.549 & 0.631 & 0.316 & 0.074 & 0.912 & 0.999 & & \\
& Temperature & 0.909 & 0.925 & 0.926 & 0.728 & 0.950 & 1.000 & & \\
& GCS         & 0.491 & 0.695 & 0.387 & 0.302 & 0.748 & 0.998 & & \\
& PaO2/FiO2   & 0.413 & 0.258 & 0.439 & 0.301 & 0.271 & 0.998 & & \\
& BUN         & 0.880 & 0.879 & 0.852 & 0.676 & 0.847 & 0.999 & & \\
& Urine       & 0.119 & 0.359 & 0.191 & 0.083 & 0.353 & 0.974 & & \\
& Sodium      & 0.939 & 0.892 & 0.947 & 0.591 & 0.856 & 1.000 & & \\
& Potassium   & 0.672 & 0.710 & 0.680 & 0.481 & 0.769 & 0.998 & & \\
& Bicarbonate & 0.859 & 0.836 & 0.823 & 0.628 & 0.855 & 0.999 & & \\
& WBC         & 0.777 & 0.655 & 0.780 & 0.606 & 0.661 & 0.997 & & \\
& Bilirubin   & 0.767 & 0.400 & 0.882 & 0.689 & 0.410 & 0.999 & & \\
\midrule
\multirow{16}{*}{\begin{sideways}SOFA current\end{sideways}} 
& GCS-eye        & 0.855 & 0.982 & 0.764 & 0.608 & 0.984 & 0.998 &  &  \\
& GCS-motor      & 0.914 & 0.980 & 0.733 & 0.616 & 0.983 & 0.998 &  &  \\
& GCS-verbal     & 0.947 & 0.985 & 0.750 & 0.635 & 0.985 & 0.999 &  &  \\
& MAP            & 0.036 & 0.219 & 0.046 & 0.089 & 0.316 & 0.993 &  &  \\
& Dopamine       & 0.875 & 0.952 & 0.754 & 0.078 & 0.950 & 0.972 &  &  \\
& Dobutamine     & 0.905 & 0.978 & 0.785 & 0.083 & 0.977 & 0.993 &  &  \\
& Epinephrine    & 0.891 & 0.965 & 0.773 & 0.082 & 0.964 & 0.979 &  &  \\
& Norepinephrine & 0.809 & 0.875 & 0.692 & 0.076 & 0.874 & 0.909 &  &  \\
& Weight         & 0.851 & 0.765 & 0.722 & 0.280 & 0.916 & 0.999 &  &  \\
& PaO2/FiO2      & 0.140 & 0.559 & 0.090 & 0.059 & 0.562 & 0.998 &  &  \\
& PaO2           & 0.533 & 0.563 & 0.352 & 0.288 & 0.574 & 0.998 &  &  \\
& FiO2           & 0.881 & 0.711 & 0.721 & 0.479 & 0.779 & 1.000 &  &  \\
& Platelet       & 0.922 & 0.966 & 0.676 & 0.664 & 0.959 & 0.997 &  &  \\
& Bilirubin      & 0.299 & 0.275 & 0.262 & 0.177 & 0.289 & 0.999 &  &  \\
& Urine          & 0.102 & 0.262 & 0.095 & 0.037 & 0.281 & 0.966 &  &  \\
& Creatinine     & 0.887 & 0.970 & 0.666 & 0.408 & 0.956 & 0.998 &  &  \\
\midrule
\multirow{16}{*}{\begin{sideways}SOFA future\end{sideways}} 
& GCS-eye        & 0.679 & 0.710 & 0.525 & 0.357 & 0.645 & 0.712 & 0.743 & 0.789 \\
& GCS-motor      & 0.783 & 0.805 & 0.620 & 0.398 & 0.746 & 0.811 & 0.832 & 0.869 \\
& GCS-verbal     & 0.806 & 0.840 & 0.648 & 0.434 & 0.786 & 0.848 & 0.871 & 0.895 \\
& MAP            & 0.110 & 0.197 & 0.125 & 0.064 & 0.219 & 0.274 & 0.289 & 0.294 \\
& Dopamine       & 0.917 & 0.966 & 0.766 & 0.386 & 0.936 & 0.978 & 0.987 & 0.988 \\
& Dobutamine     & 0.937 & 0.982 & 0.787 & 0.388 & 0.955 & 0.994 & 0.993 & 0.993 \\
& Epinephrine    & 0.933 & 0.972 & 0.779 & 0.381 & 0.945 & 0.989 & 0.989 & 0.989 \\
& Norepinephrine & 0.861 & 0.887 & 0.710 & 0.350 & 0.849 & 0.913 & 0.914 & 0.913 \\
& Weight         & 0.851 & 0.765 & 0.722 & 0.280 & 0.916 & 0.915 & 0.916 & 0.914 \\
& PaO2/FiO2      & 0.073 & 0.156 & 0.053 & 0.038 & 0.198 & 0.565 & 0.584 & 0.596 \\
& PaO2           & 0.181 & 0.179 & 0.148 & 0.098 & 0.227 & 0.616 & 0.659 & 0.715 \\
& FiO2           & 0.738 & 0.538 & 0.613 & 0.379 & 0.620 & 0.821 & 0.840 & 0.867 \\
& Platelet       & 0.283 & 0.300 & 0.236 & 0.166 & 0.334 & 0.364 & 0.383 & 0.396 \\
& Bilirubin      & 0.180 & 0.166 & 0.169 & 0.115 & 0.175 & 0.811 & 0.832 & 0.866 \\
& Urine          & 0.064 & 0.077 & 0.069 & 0.023 & 0.079 & 0.133 & 0.152 & 0.183 \\
& Creatinine     & 0.360 & 0.347 & 0.295 & 0.161 & 0.392 & 0.395 & 0.424 & 0.455 \\
\bottomrule
    \end{tabular}
}
\caption{Value correctness of predictions compared to real-world measurements.}
    \label{tab:results_value_clinical}
\end{table*}

\section{Experimental Results}
\label{sec:results}

\paragraph{LLMs Shine in Deductive Inference According to Complex Consensus Rules}

Our first research question asks whether LLMs can learn the deductive inference rules underlying medical consensus definitions by finetuning on verbalized instantiations of the inference process to patient data. Table \ref{tab:results_inference} shows that all finetuned models, from basic finetuning with text encoding to pipeline and multimodal approaches, achieve nearly perfect derivation correctness. This includes extracting worst values, mapping clinical variables to non-linear SAPS-II and SOFA step functions for the first 24 hours (first and second block), mapping forecasted clinical variables to future SOFA scores (third block), and computing a Sepsis label (last row). In contrast, one-shot learning does not perform nearly as good, even for models that are specialized for reasoning (deepseek, qwen), pretrained on medical guidelines (me-llama), or are significantly larger (one-shot-70B). Two recent models reinforce this (Appendix~\ref{appendix:further-llms}): HuatuoGPT-o1-8B reaches only Sepsis-3 accuracy $0.765$/F1 $0.204$, below the one-shot LLaMA-3-8B baseline ($0.834$/$0.231$), and Olmo-3-7B-Instruct fails to follow the template at all. 
Forced derivation correctness (in brackets in the third block) shows similar relations using correct histories. Again, all one-shot learners underperform compared to finetuning.\footnote{Note that enforcing correct histories can lower results, e.g., for $\text{SOFA}_\text{diff}$, since predicting a constant difference from wrongly predicted constant current and future SOFA values is a correct derivation.}  Overall, this shows that learning medical inference rules needs more supervision than one-shot learning, despite including the explicit inference rules and an example instantiation to patient data in the one-shot prompts. Most striking is the non-linear SAPS-II step, where finetuning reproduces the logit/sigmoid almost perfectly while one-shot baselines collapse even with the formula in the prompt (Appendix~\ref{appendix:further-analyses}).

\begin{table*}[t!]
    \centering
\resizebox{0.7\textwidth}{!}{
    \begin{tabular}{lllllll|lll}
    \toprule
 & variable & one-shot & one-shot-70B & me-llama & deepseek & qwen3.6-27B & finetuned & pipeline & multimodal \\
\midrule
\multirow{16}{*}{\begin{sideways}SAPS-II current\end{sideways}}
& S$_\text{Age}$          & 0.138 & 0.200 & 0.133 & 0.090 & 0.171 & 1.000 & & \\
& S$_\text{HR}$           & 0.270 & 0.506 & 0.393 & 0.333 & 0.675 & 0.999 & & \\
& S$_\text{SBP}$          & 0.244 & 0.482 & 0.299 & 0.361 & 0.853 & 0.999 & & \\
& S$_\text{Temperature}$  & 0.779 & 0.759 & 0.928 & 0.739 & 0.927 & 0.999 & & \\
& S$_\text{GCS}$          & 0.487 & 0.710 & 0.486 & 0.405 & 0.775 & 0.999 & & \\
& S$_\text{PaO2/FiO2}$    & 0.432 & 0.251 & 0.573 & 0.507 & 0.203 & 0.999 & & \\
& S$_\text{BUN}$          & 0.592 & 0.751 & 0.670 & 0.534 & 0.825 & 1.000 & & \\
& S$_\text{Urine}$        & 0.225 & 0.553 & 0.181 & 0.146 & 0.837 & 0.995 & & \\
& S$_\text{Sodium}$       & 0.806 & 0.578 & 0.858 & 0.686 & 0.726 & 1.000 & & \\
& S$_\text{Potassium}$    & 0.820 & 0.817 & 0.838 & 0.668 & 0.842 & 1.000 & & \\
& S$_\text{Bicarbonate}$  & 0.837 & 0.792 & 0.844 & 0.670 & 0.764 & 1.000 & & \\
& S$_\text{WBC}$          & 0.863 & 0.771 & 0.887 & 0.709 & 0.840 & 1.000 & & \\
& S$_\text{Bilirubin}$    & 0.900 & 0.429 & 0.924 & 0.733 & 0.502 & 1.000 & & \\
& Total SAPS-II          & 0.077 & 0.136 & 0.068 & 0.017 & 0.136 & 0.993 & & \\
& Logit             & 0.032 & 0.035 & 0.027 & 0.022 & 0.051 & 0.991 & & \\
& Probability       & 0.031 & 0.044 & 0.024 & 0.015 & 0.061 & 0.992 & & \\
\midrule
\multirow{7}{*}{\begin{sideways}SOFA current\end{sideways}} 
& S$_\text{CNS}$       & 0.633 & 0.794 & 0.478 & 0.239 & 0.982 & 0.998 &  &  \\
& S$_\text{cardio}$    & 0.405 & 0.600 & 0.249 & 0.273 & 0.688 & 0.985 &  &  \\
& S$_\text{resp}$      & 0.223 & 0.411 & 0.172 & 0.128 & 0.652 & 0.996 &  &  \\
& S$_\text{coag}$      & 0.614 & 0.761 & 0.497 & 0.428 & 0.868 & 0.999 &  &  \\
& S$_\text{liver}$     & 0.492 & 0.809 & 0.502 & 0.208 & 0.900 & 1.000 &  &  \\
& S$_\text{renal}$     & 0.540 & 0.735 & 0.412 & 0.303 & 0.833 & 0.997 &  &  \\
& $\text{SOFA}_{1:24}$ & 0.111 & 0.236 & 0.076 & 0.044 & 0.441 & 0.979 &  &  \\
\midrule
\multirow{8}{*}{\begin{sideways}SOFA future\end{sideways}} 
& S$_\text{CNS}$            & 0.532 & 0.625 & 0.408 & 0.231 & 0.635 & 0.701 & 0.732 & 0.760 \\
& S$_\text{cardio}$         & 0.426 & 0.527 & 0.295 & 0.195 & 0.589 & 0.700 & 0.738 & 0.745 \\
& S$_\text{resp}$           & 0.201 & 0.348 & 0.163 & 0.082 & 0.576 & 0.746 & 0.754 & 0.780 \\
& S$_\text{coag}$           & 0.618 & 0.693 & 0.483 & 0.278 & 0.747 & 0.839 & 0.823 & 0.834 \\
& S$_\text{liver}$          & 0.462 & 0.783 & 0.570 & 0.148 & 0.839 & 0.945 & 0.945 & 0.945 \\
& S$_\text{renal}$          & 0.482 & 0.620 & 0.365 & 0.211 & 0.658 & 0.553 & 0.451 & 0.473 \\
& $\text{SOFA}_{25:48}$     & 0.117 & 0.176 & 0.088 & 0.059 & 0.219 & 0.203 & 0.220 & 0.243 \\
& $\text{SOFA}_\text{diff}$ & 0.139 & 0.167 & 0.128 & 0.127 & 0.209 & 0.151 & 0.147 & 0.163 \\
\midrule
\multirow{4}{*}{\begin{sideways}SEPSIS-3\end{sideways}}
& Accuracy    & 0.834 & 0.857 & 0.763 & 0.692 & 0.891 & 0.886 & 0.877 & 0.882 \\
& Specificity & 0.876 & 0.915 & 0.788 & 0.739 & 0.953 & 0.936 & 0.920 & 0.922 \\
& Sensitivity & 0.332 & 0.118 & 0.455 & 0.091 & 0.081 & 0.263 & 0.336 & 0.386 \\
& F1          & 0.231 & 0.108 & 0.220 & 0.042 & 0.101 & 0.254 & 0.272 & 0.309 \\
\bottomrule
    \end{tabular}
}
    \caption{Value correctness of predicted SOFA, SEPSIS and SAPS-II scores compared to derivations from real-world clinical measurements.} 
    \label{tab:results_value_sofa}
\end{table*}

\begin{table}[t!]
    \centering
    \resizebox{\linewidth}{!}{
    \begin{tabular}{lllll}
     \toprule
variable & \% changes ID & \% changes OOD & ID score & OOD score\\
\midrule
$\text{SOFA}_{1:24}$ & 0.404  & 0.424 & 1.000 & 1.000\\ 
$\text{SOFA}_{25:48}$& 0.433 & 0.452 & 1.000 & 1.000\\
\bottomrule
    \end{tabular}
    }
    \caption{Derivation correctness for in- and out-of-distribution synthetic exceptions to SOFA inference rules.}
    \label{tab:synt_icd_results}
\end{table}

\begin{table}[t!]
    \centering
    \resizebox{\linewidth}{!}{
    \begin{tabular}{lllllll}
    \toprule
    variable & one-shot & one-shot-70B & me-llama & deepseek & qwen3.6-27B & finetuned \\
    \midrule
    S$_\text{CNS}$       & 0.727 & 0.773 & 0.603 & 0.385 & 0.974 & 0.998 \\
    S$_\text{cardio}$    & 0.380 & 0.531 & 0.311 & 0.306 & 0.653 & 0.984 \\
    S$_\text{resp}$      & 0.605 & 0.377 & 0.505 & 0.395 & 0.700 & 0.995 \\
    S$_\text{coag}$      & 0.573 & 0.658 & 0.554 & 0.280 & 0.799 & 0.999 \\
    S$_\text{liver}$     & 0.625 & 0.843 & 0.858 & 0.573 & 0.923 & 1.000 \\
    S$_\text{renal}$     & 0.463 & 0.651 & 0.577 & 0.274 & 0.827 & 0.998 \\
    $\text{SOFA}_{1:24}$ & 0.152 & 0.145 & 0.120 & 0.052 & 0.407 & 0.979 \\
    \bottomrule
    \end{tabular}
    }
    \caption{Value correctness of current SOFA scores with data-driven exceptions to SOFA rules (future scores in Table~\ref{tab:results_value_sofa_exception}).}
    \label{tab:results_exc_main}
\end{table}

\paragraph{LLMs Reliably Learn Selected Exceptions to Consensus Rules}
The next question we ask is whether LLMs can learn exceptions to the inference rules that arise because of special treatments of chronic comorbidities or interventional therapies for particular patient cohorts.  We used the exceptions to the SOFA medical guidelines shown in Table \ref{tab:icd9_mappings} in Appendix \ref{appendix:exc}, affecting renal, liver, CNS, and coagulation SOFA systems. Table \ref{tab:results_inference_exception} shows that finetuning on data including exceptions to the rule yields near-perfect derivation correctness results as on data without exceptions. This capacity of detecting exceptional rule patterns does not emerge in one-shot learning. Perfect derivation correctness is also achieved under covariate shift in a controlled setup with synthesized exceptions (see Table \ref{tab:synt_icd_results}; setup in Appendix \ref{appendix:exc}).
Since exceptions are taught to the models only through examples -- either one-shot demonstration examples or finetuning data -- but never stated as abstract rules, finetuning succeeds in mapping premises like ICD codes or therapies to conclusions resulting in changes in the SOFA scores of the corresponding organ systems, but one-shot learning from a single demonstration fails in this associative learning task.  This holds under covariate shift scenarios as well: with held-out ICD codes from known ICD classes seen only at test time, derivation correctness stays at $1.000$ (Table~\ref{tab:synt_icd_results}), indicating the model learns the exception \emph{pattern} within this synthetic schema, rather than memorizing premise tokens. Scores with exceptions also hold up against real-world values: for the first 24 hours, finetuning reaches value correctness of $0.979$--$1.000$, while the best baseline stays at $0.407$ on total SOFA (Table~\ref{tab:results_exc_main}).

\paragraph{On the Evaluated Tasks, the Bottleneck Lies in Inductive Inference}

Our final evaluation compares the results of partial inference steps to real-world clinical measurements and to SAPS-II, SOFA and Sepsis scores derived from these. As shown in the first and second block of Table \ref{tab:results_value_clinical}, the numerical values that a finetuned LLM decoder extracts from current time series are nearly perfect for clinical variables relevant for SAPS-II and SOFA. The third block of Table \ref{tab:results_value_clinical}  reveals the  bottleneck of teaching medical inference rules to LLMs: Correctness of predicting values of 131 clinical variables 24 hours into the future drops from the high nineties to values in the tens and twenties, especially for sparsely and irregularly observed lab values. As shown in Table \ref{tab:results_value_sofa}, the same is true for value correctness of predicted scores: The numerical values produced by mapping clinical measurements in the first 24 hours to SAPS-II and SOFA scores are nearly perfect, however, mapping values of forecasted clinical variables to step functions severely impacts correctness, especially on metrics like F1 that are sensitive to imbalanced data. One-shot learning is impacted even more, while finetuning results can be improved by moving from a text-based encoding to coupling dedicated TSF encoders with the LLM. 

This improvement is not uniform: future S$_\text{renal}$ drops from $0.553$ (finetuned) to $0.451$ (pipeline) and $0.473$ (multimodal), all below qwen3.6-27B ($0.658$) and one-shot-70B ($0.620$), and on $\text{SOFA}_\text{diff}$ all finetuned variants ($0.147$--$0.163$) stay below qwen3.6-27B ($0.209$) and one-shot-70B ($0.167$). This is likely because renal SOFA depends on urine output, a daily sum that is harder to forecast than minima or maxima and is sparsely and irregularly measured, which is reflected in its low value correctness ($0.133$--$0.183$). At a positive class prevalence of $7.3\%$, a majority baseline yields accuracy $0.927$ and F1 $0$, so F1 is our main metric; finetuned models are not best on accuracy (qwen3.6-27B $0.891$) or sensitivity (me-llama $0.455$). Similar relations are obtained when including exceptional inference rules (see Appendix \ref{appendix:results_exc}).

We emphasize that the evaluation of \emph{value correctness} against real measurements shows that learning rules and exceptions goes beyond mere template reproduction, otherwise the gap between current and future values would not exist. Furthermore, we note that the Sepsis-3 F1 stays modest (multimodal $0.309$). However, answer-only supervision of the multimodal model with a direct prediction head (F1 $0.277$, AUROC $0.820$) and XGBoost (F1 $0.300$--$0.332$, AUROC $0.831$--$0.856$) stay in a similar range on the same 24-hour window, which suggests that the long prediction horizon limits performance, rather than the guideline-based pipeline (Appendix~\ref{appendix:further-analyses}).

\section{Conclusion and Limitations}
\label{sec:disc}

LLMs have been shown to excel at various health-related tasks, however, as stated by \cite{Adler-Milstein:21}, diagnostic AI has not realized the potential to reduce the cognitive demand and associated diagnostic errors of clinicians, due to a focus on predictive accuracy of final diagnostic labels. Furthermore, \cite{AngusBindman:22} advocate to learn patterns of clinical data that are yet unrecognized but still predictive of sepsis.
Our approach can be seen as a first step towards these goals by putting a focus on teaching LLMs inference rules that support clinicians, and to learn exceptions to the rules that capture clinical patterns that are not yet anchored in medical consensus. 
Given the ease of automatically generating in-distribution finetuning data and the nearly perfect results of deductive inference of finetuned models, we conjecture that, on the evaluated tasks, the bottleneck of medical reasoning LLMs is not out-of-distribution rule application, but the orthogonal problem of generalization into the future for inductive inference. Future work shall further address the forecasting bottleneck and conduct a deeper investigation of learning exceptions to consensus rules. 

The limitations of our current work are as follows: Our three guidelines are all structured, numerical ICU scores for diagnosis and prognosis, chosen so the gold standard is automatically computable, so our results do not extend to guidelines in general; generalization to qualitative guideline families, e.g., \ DSM-5 criteria \citep{DSMV:13}, is future work. 

All experiments use a single dataset (MIMIC-III); cross-cohort validation on MIMIC-IV/eICU remains to be done. Exceptions are hand-curated or synthetic; discovering new exceptions and routing them to unseen patients is future work. 

Parsing may penalize models with other output formats, and robustness to perturbed verbalizations is untested. 

Finetuned models see 15,000 examples and baselines one, so our comparison shows the value of task-specific supervision; finetuned comparators, few-shot prompting, and explicit exception rules remain to be evaluated. 

Furthermore, we did not evaluate reinforcement learning or other methods targeting distribution shift \citep{ChuETAL:25,DeepSeek:25,WenETAL:25}, so we cannot rule out that they would help. 

Lastly, unlike a rule engine, the LLM has a real training cost (Appendix~\ref{appendix:hyperparameters}).

\acks{The first author of this work was funded by an industry transfer grant provided by the Interdisciplinary Center for Scientific Computing (IWR) at Heidelberg University. 

The authors gratefully acknowledge the computing time provided on the high-performance computer HoreKa by the National High-Performance Computing Center at KIT (NHR@KIT). This center is jointly supported by the Federal Ministry of Research, Technology and Space and the Ministry of Science, Research and the Arts of Baden-Württemberg, as part of the National High-Performance Computing (NHR) joint funding program. HoreKa is partly funded by the German Research Foundation (DFG).

The authors furthermore acknowledge support by the state of Baden-Württemberg through bwHPC (bwUniCluster, bwForCluster Helix) and the German Research Foundation (DFG)
through grant INST 35/1597-1 FUGG (bwForCluster Helix).}

\bibliography{ref}

\newpage
\appendix
\onecolumn

\section{Supplemental Material}

\subsection{Data}
\label{appendix:data}

In our experiments, we use electronic health records (EHRs) from the MIMIC-III data \citep{JohnsonETAL:16}. After filtering for patients with an ICU stay of at least 24 hours with reported gender and age of at least 18 years, our dataset contained 44,858 ICU stays with 56 million datapoints, which are two-day sliding window time series. We split the data strictly per patient, with no patient overlap between partitions, into finetuning (28,708 stays), development (7,270 stays), and testing (8,880 stays). For computational reasons, we further subsampled 15,000 datapoints for finetuning, and 3,000 datapoints for development and testing, respectively. 
The resulting percentage of positive Sepsis-3 cases in the test set is 7.3\%, so a majority baseline that always predicts the negative class obtains accuracy 0.927, specificity 1.000, sensitivity 0.000, and F1 0.000.
As features, we considered 131 clinical variables and the demographic variables gender and age (see Appendix \ref{appendix:features}). This selection comprises all vital signs and laboratory values used in the PhysioNet challenge for early prediction of sepsis \citep{ReynaETAL:19}, together with information on suspected infection\footnote{The onset of suspicion of infection is thus given prior to the onset of salient organ dysfunction in our approach, corresponding to labeling scheme H1 of \cite{CohenETAL:24}. This is consistent with other work where suspected infection is included in the list of input features \citep{NematiETAL:18}.}. 
The time series of clinical observations in the data were split into a 24 hour observation window, followed by a 24 hour prediction window. During training and testing, we use a sliding window of 24 hours so that full admission days are given as input and output. Using the consensus definitions, SOFA and SAPS-II scores were calculated for the observation window (current). For Sepsis-3, values of clinical variables and SOFA score were forecasted for the prediction window (future), and final labels were calculated deterministically for the given time series data.  In the calculation of the ground truth, missing values are being carried forward from the previous day only. 

\newpage 

\subsection{Evaluation Metrics}
\label{appendix:evaluation-metrics}

\begin{figure*}[h!]
    \centering
    \includegraphics[width=0.95\textwidth]{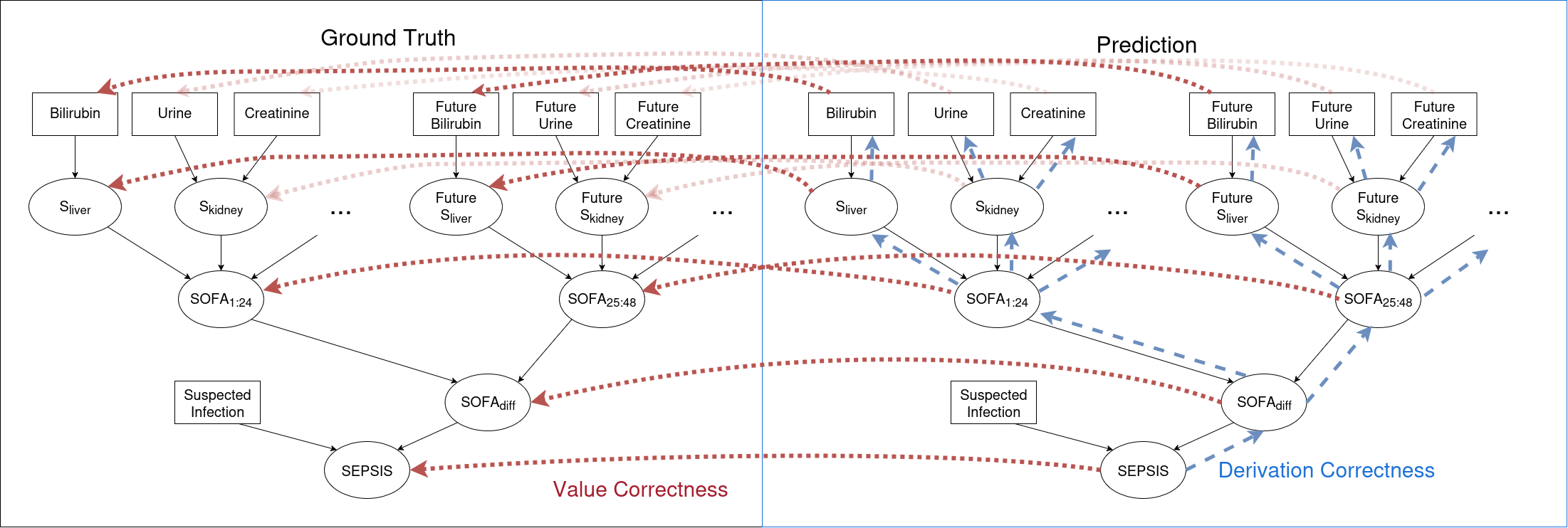}
    \caption{\textbf{Derivation correctness} (dashed blue arrows) checks for the predicted inference graph whether each child node (conclusion) follows from the parent node (premise) according to the consensus rule. \textbf{Value correctness} (dotted red arrows) maps each node in the predicted inference graph to its corresponding node in the ground truth graph, consisting of real-world clinical measurements for first and second 24 hours, and deterministic calculation of SOFA and SEPSIS on these values. Evaluation for SAPS-II is done analogously.}
    \label{fig:evaluation-metrics}
\end{figure*}

\paragraph{Statistical significance.} We use an approximate randomization test, also known as a permutation test \citep{Noreen:89}, with stratified shuffling on the patient level. For two models, we compute the absolute difference of their scores on the test set. In each of 10,000 trials, the predictions of the two models are swapped with probability $0.5$ and the absolute difference is recomputed. The p-value is the fraction of trials in which this difference is greater than or equal to the observed one, and we call a difference significant if $p<0.05$. For Sepsis-3 F1, this yields $p=0.043$ for finetuned against me-llama and $p=0.048$ for finetuned against one-shot. For intermediate scores such as S$_\text{CNS}$ in Table~\ref{tab:results_value_sofa}, we replace F1 with our matching function; here, none of the 10,000 trials reached the observed difference. All pairwise comparisons of finetuning against one-shot learning yield $p<0.05$.

\newpage

\subsection{SOFA Score}
\label{appendix:sofa}

\begin{table*}[h!]
	\centering
	\small
	        \resizebox{0.99\textwidth}{!}{
\begin{tabular}{rp{0.8cm}p{4.5cm}p{1.5cm}p{1.5cm}p{1.5cm}p{1.5cm}}
    \toprule
          & CNS         & Cardiovascular                                                                                                                        & Respiratory          & Coagulation              & Liver             & Renal                                  \\
          \cmidrule{2-7}
   \rotatebox{90}{Score\hspace*{-1em}} & GCS         & MAP  \newline or vasopressors                                                                      & PaO2/FiO2 \newline(mmHg)       & Platelets \newline ($\times10^3$/$\mu$l) & Bilirubin \newline(mg/dl) & Creatinine \newline(mg/dl)  or UO             \\ \cmidrule{2-7}
    +0    & 15          & MAP $\geq$ 70 mmHg                                                                                                                    & $\geq$ 400           & $\geq$ 150                 & $<$ 1.2     & $<$ 1.2                                       \\
    +1    & 13-14       & MAP $<$ 70 mmHg                                                                                                                 & $<$ 400        & $<$ 150              & 1.2-1.9           & 1.2-1.9                                             \\
    +2    & 10-12       & dopamine $\leq$ 5 $\mu$g/kg/min \newline OR dobutamine (any dose)                                                                      & $<$ 300        & $<$ 100              & 2.0-5.9           & 2.0-3.4                                             \\
    +3    & 6-9         & dopamine $>$ 5 $\mu$g/kg/min \newline OR epinephrine $\leq$ 0.1 $\mu$g/kg/min \newline OR norepinephrine $\leq$ 0.1 $\mu$g/kg/min              & $<$ 200 \newline AND MV & $<$ 50               & 6.0-11.9          & 3.5-4.9 \newline OR $<$ 500 ml/day                   \\
    +4    & $<$ 6 & dopamine $>$ 15 $\mu$g/kg/min \newline OR epinephrine $>$ 0.1 $\mu$g/kg/min \newline OR norepinephrine $>$ 0.1 $\mu$g/kg/min & $<$ 100 \newline AND MV & $<$ 20               & $>$ 12.0 & $>$ 5.0 \newline OR $<$ 200 ml/day\\\bottomrule
\end{tabular}
}
\caption{SOFA score for six organ systems, calculated by thresholding corresponding clinical measurements \citep{VincentETAL:96}. In our setting, we had to recalculate MAP (mean arterial pressure) from SBP and DBP (systolic and diastolic blood pressure), the Horowitz coefficient from PaO2 and FiO2, and had no knowledge about the kind of mechanical ventilation. If no value for calculation in a SOFA subsystem was available, we took a value of 0. Abbreviations: CNS = Central nervous system; GCS = Glasgow Coma Scale; MV = mechanically ventilated including CPAP; MAP = mean arterial pressure, UO = Urine output.}
    \label{tab:sofa}
\end{table*}

\subsection{SAPS-II Score}
\label{appendix:saps}
On top of the thresholded SAPS-II score, a logit function is computed and converted to a probability of hospital mortality via a sigmoid:
\begin{align*}
\text{logit} & = \beta_0 + \beta_1(\text{SAPS II score}) 
+ \beta_2 \left[\ln(\text{SAPS II score} + 1)\right] \\
& = -7.7631 + 0.0737(\text{SAPS II score})
+ 0.9971 \left[\ln(\text{SAPS II score} + 1)\right].
\end{align*}
The probability of hospital mortality is computed as 
$
\text{Pr}(y=1 \mid \text{logit}) =
\frac{e^{\text{logit}}}{1 + e^{\text{logit}}}.
$

\begin{landscape}
\begin{table*}[h!]
\small
	\begin{threeparttable}[b]
	\centering
	\small
	\begin{tabular}{lcccccccccccc}
		\toprule
		Score & HR\tnote{1} & SBP\tnote{1} &  Temp\tnote{2}  & GCS\tnote{3} & PaO2/Fi02\tnote{3} & BUN\tnote{2} & Urine\tnote{4} & Sodium\tnote{1} & Potassium\tnote{1} & Bicarbonate\tnote{3} & WBC\tnote{1} & Bilirubin\tnote{2} \\
		      &    [1/s]    &    [mmHG]    & [\textdegree C] &              &     [mmHg/\%]      &   [mg/dl]    &    [ml/24h]    &    [mmol/L]     &      [mEq/L]       &       [mEq/L]        &  [1000/mm³]  &      [mg/dL]       \\ \midrule
		+0    &   70-119    &   100-199    &      else       &    14-15     &       not MV       &     $<$28      &   $\geq$1000   &     125-144     &      3.0-4.9       &       $\geq$20       &   1.0-19.9   &       $<$4.0       \\
		+1    &             &              &                 &              &                    &              &                &    $\geq$145    &                    &                      &              &                    \\
		+2    &    40-69    &    $>$199    &                 &              &                    &              &                &                 &                    &                      &              &                    \\
		+3    &             &              &    $\geq$39     &              &                    &              &                &                 &                    &        15-19         &   $\geq$20   &                    \\
		+4    &   120-159   &              &                 &              &                    &              &    500-599     &                 &                    &                      &              &      4.0-5.9       \\
		+5    &             &    70-99     &                 &    11-13     &                    &              &                &     $<$125      &  $<$3 or $\geq$5   &                      &              &                    \\
		+6    &             &              &                 &              &       $>$199       &    28-83     &                &                 &                    &        $<$15         &              &                    \\
		+7    &   $>$159    &              &                 &     9-10     &                    &              &                &                 &                    &                      &              &                    \\
		+9    &             &              &                 &              &      100-199       &              &                &                 &                    &                      &              &      $\geq$6       \\
		+10   &             &              &                 &              &                    &    $>$83     &                &                 &                    &                      &              &                    \\
		+11   &    $<$40    &              &                 &              &       $<$100       &              &     $<$500     &                 &                    &                      &              &                    \\
		+12   &             &              &                 &              &                    &              &                &                 &                    &                      &    $<$1.0    &                    \\
		+13   &             &    $<$70     &                 &     6-8      &                    &              &                &                 &                    &                      &              &                    \\
		+26   &             &              &                 &     $<$6     &                    &              &                &                 &                    &                      &              &                    \\ \bottomrule
		      &             &              &                 &              &
	\end{tabular}
	\begin{tablenotes}
		\item [1] worst (min or max) value in 24h
		\item [2] largest (max) value in 24h
		\item [3] smallest (min) value in 24h
		\item [4] total (sum) value in 24h
	\end{tablenotes}	
\end{threeparttable}
\caption{List of  non-static components of SAPS-II score \citep{LeGallETAL:93}. 
			The total score is computed by summing the individual contributions. 
			Abbreviations: HR = heart rate; SBP = systolic blood pressure;  GCS = Glasgow Coma Scale; BUN = blood urea nitrogen; WBC = white blood cell count, MV = mechanically ventilated including CPAP.}
\label{tab:sapsii}
\end{table*}
\end{landscape}

\newpage

\subsection{Clinical Features}
\label{appendix:features}

\begin{table*}[h!]
\centering
    \resizebox{0.99\textwidth}{!}{
\begin{tabular}{llll}

\toprule
ALP                  & Epinephrine*          & LDH                       & Packed RBC           \\
ALT                  & Famotidine            & Lactate                   & Pantoprazole         \\
AST                  & Fentanyl              & Lactated Ringers          & Phosphate            \\
Albumin              & FiO2*                 & Levofloxacin              & Piggyback            \\
Albumin 25\%         & Fiber                 & Lorazepam                 & Piperacillin         \\
Albumin 5\%          & Free Water            & Lymphocytes               & Platelet Count*      \\
Amiodarone           & Fresh Frozen Plasma   & Lymphocytes (Absolute)    & Potassium            \\
Anion Gap            & Furosemide            & MBP                       & Pre-admission Intake \\
BUN                  & GCS\_eye*             & MCH                       & Pre-admission Output \\
Base Excess          & GCS\_motor*           & MCHC                      & Propofol             \\
Basophils            & GCS\_verbal*          & MCV                       & RBC                  \\
Bicarbonate          & GT Flush              & Magnesium                 & RDW                  \\
Bilirubin (Direct)   & Gastric               & Magnesium Sulfate (Bolus) & RR                   \\
Bilirubin (Indirect) & Gastric Meds          & Magnesium Sulphate        & Residual             \\
Bilirubin (Total)*   & Glucose (Blood)       & Mechanically ventilated   & SBP*                 \\
CRR                  & Glucose (Serum)       & Metoprolol                & SG Urine             \\
Calcium Free         & Glucose (Whole Blood) & Midazolam                 & Sodium               \\
Calcium Gluconate    & HR                    & Milrinone                 & Solution             \\
Calcium Total        & Half Normal Saline    & Monocytes                 & Sterile Water        \\
Cefazolin            & Hct                   & Morphine Sulfate          & Stool                \\
Chest Tube           & Heparin               & Neosynephrine             & TPN                  \\
Chloride             & Hgb                   & Neutrophils               & Temperature          \\
Colloid              & Hydralazine           & Nitroglycerine            & Total CO2            \\
Creatinine Blood*    & Hydromorphone         & Nitroprusside             & Ultrafiltrate        \\
Creatinine Urine     & INR                   & Norepinephrine*           & Urine*               \\
D5W                  & Insulin Humalog       & Normal Saline             & Vancomycin           \\
DBP*                 & Insulin NPH           & O2 Saturation             & Vasopressin          \\
Dextrose Other       & Insulin Regular       & OR/PACU Crystalloid       & WBC                  \\
Dobutamine*          & Insulin largine       & PCO2                      & Weight               \\
Dopamine*            & Intubated             & PO intake                 & pH Blood             \\
EBL                  & Jackson-Pratt         & PaO2*                      & pH Urine             \\
Emesis               & KCl                   & PT                        &                      \\
Eoisinophils         & KCl (Bolus)           & PTT                       &          \\
\bottomrule
\end{tabular}
}
\caption{Feature list for MIMIC-III: Besides the following 131 dynamic variables, only age and gender were extracted. The 15 variables marked with an asterisk are directly used for calculating the SOFA score.}
\label{tab:features_mimic}
\end{table*}

\newpage

\subsection{Exceptions to SOFA Medical Guidelines}
\label{appendix:exc}

\begin{table*}[h!]
\centering
\resizebox{0.99\textwidth}{!}{
\begin{tabular}{p{4cm}p{\dimexpr\linewidth-4.5cm\relax}p{3cm}}
\toprule
\textbf{Category} & \textbf{Premise} & 
\textbf{Conclusion}\\
\midrule
Chronic Kidney Disease Stage 3 (CKD3) & 585.3, 585.9 & Renal SOFA = 2 \\
CKD4 & 585.4 & Renal SOFA = 3\\
CKD5 & 585.5, 585.6, 403.01, 403.11, 403.91, 404.02, 404.03, 404.13, 404.93 & Renal SOFA = 4 \\
Leukemia or Lymphoma & 201.0, 201.1, 201.6, 201.7, 201.2, 201.9, 202.0, 202.8, 200.7, 200.1, 200.2, 202.1, 202.2, 202.7, 200.6, 200.3, 203.8, 203.0, 203.1, 238.6, 204.0, 204.1, 204.8, 202.4, 204.9, 205.0, 205.1, 205.3, 205.9, 206.0, 206.1, 206.2, 206.9, 207.0, 207.2, 207.8, 208.0, 208.1, 208.9, 202.5, 202.6, 202.9, 200.0, 200.9, 201.4, 201.5, 203.00, 203.01, 203.02, 203.10, 203.11, 203.12, 204.00, 204.01, 204.02, 204.10, 204.11, 204.12, 205.00, 205.01, 205.02, 205.10, 205.11, 205.12, 206.00, 206.01, 206.10, 206.11, 207.00, 207.01, 207.20, 207.21, 208.00, 208.01, 208.10, 208.11, 208.90, 208.91 & Coag. SOFA = 4\\
Moderate-Severe Liver Disease & 456.0, 456.1, 456.21, 572.8, 572.3, 572.4 & Liver SOFA = 2\\
Dementia & 290.11, 290.0, 290.41, 290.40, 290.43, 294.10, 294.11, 290.3, 331.2, 290.42, 290.8, 290.9, 290.10, 290.12, 290.13, 290.20, 290.21 & CNS SOFA = 2\\
\midrule
Therapy & Renal Replacement Therapy (RRT) & Renal SOFA = 4 \\
\bottomrule
\end{tabular}
}
\caption{Exceptions to SOFA rules by diagnostic category (comorbidity ICD-9 code/interventional therapy).}
\label{tab:icd9_mappings}
\end{table*}

\begin{figure}[H]
    \centering
    \includegraphics[width=0.85\linewidth]{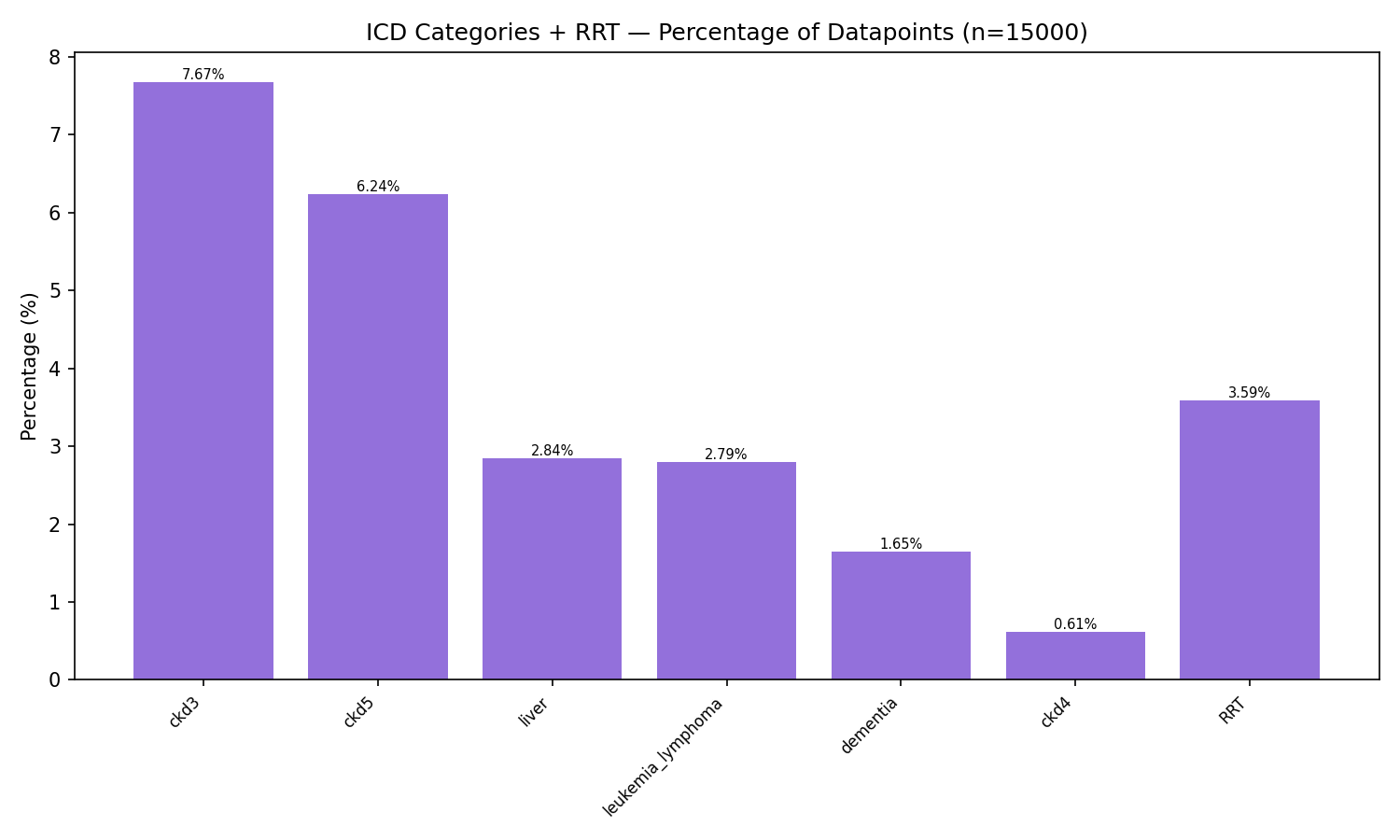}
    \caption{Percentage of diagnostic categories appearing in the train data set.}
    \label{fig:icd9_histogram}
\end{figure}

\newpage

\newpage

\begin{table*}[h!]
	\centering
    \resizebox{0.75\textwidth}{!}{
\begin{tabular}{cccccc}
    \toprule
          & lung & kidney & coagulation & liver & cardiovascular \\
          \midrule
          ID  & J40, J41, J42 & N18.9, N28 & D68.4, D68.5 & K70.0, K70.41 & I50.0, I50.9\\
          OOD & J44.9 & N19 & D68.6 & K70.3 & I50.1 \\
          \bottomrule
\end{tabular}
}
\caption{Synthetic exceptions to SOFA rules: Premises are indicated by ICD-10 codes, conclusions are increases of the corresponding SOFA scores to maximal value. During data generation, each datapoint is randomly assigned a precondition, either one of the predefined types or the no-precondition option, with equal probability. In-distribution (ID) data  were seen, out-of-distribution (OOD) data were unseen during finetuning.
}
    \label{tab:synth_icd}
\end{table*}

\begin{table*}[h!]
    \centering
    \resizebox{0.45\textwidth}{!}{
    \begin{tabular}{lll}
\toprule
ICD-10 Code & \# ID Test & \# OOD Test \\ 
\midrule
J40  &  263 & 194 \\
J41  &  239 &  173   \\
J42  &  253 & 194  \\
J44.9 &  - &  166  \\
\midrule
N18.9  &  266 & 172 \\
N19  &  - &  204  \\
N28  &  281 & 151 \\
\midrule
D68.4  &  242 &  171   \\
D68.5  &  244 &  162 \\
D68.6 & - &  176  \\
\midrule
I50.0  &  264 & 185 \\
I50.1 &  - &  193  \\
I50.9  &  229  &  162  \\
\midrule
K70.0  &  230 &  165   \\
K70.3 &  - &  162  \\
K70.41  &  215  &  182   \\
\midrule
No-Preexisting  &  274 & 189 \\
\bottomrule
    \end{tabular}
    }
    \caption{ICD-10 code frequency for synthetic exceptions in in- and out-of-distribution testsets.}
    \label{tab:synth_icd_histogram}
\end{table*}

\newpage

\subsection{Finetuning Data for Sepsis-3}
\label{appendix:verbalization}

\begin{figure*}[h!]
    \centering
     \begin{minipage}[t]{0.48\textwidth}
\scriptsize
Patient is 75.0 years old and is male. Given all the information in this text, answer the question at the end.
Here are the measurements: DBP at time -22.37: 49.0, SBP at time -22.37: 105.0, DBP at time -20.37: 52.0, GCS\_eye at time -20.37: 4.0, GCS\_motor at time -20.37: 6.0, GCS\_verbal at time -20.37: 1.0, SBP at time -20.37: 117.0, DBP at time -19.37: 56.0, FiO2 at time -19.37: 0.5, SBP at time -19.37: 127.0, DBP at time -18.37: 43.0,...
\\~\\
Now answer the following question:
The doctors suspect an infection, based on this information and the other information in this text, will the patient be classified as septic tomorrow?
\vspace{0.5em}
\hrule
\vspace{0.5em}
First we need to calculate the SOFA scores given the extracted values. The SOFA scores for the current time are the following:
The minimum value of GCS\_eye is 4.0, GCS\_motor is 6.0 and GCS\_verbal is 1.0, this produces the sum 11.0 and means the CNS SOFA is 2.
Because minimum MAP is 55.333, max Dopamine is 0, max Dobutamine is 0, max Epinephrine is 0 and max Norepinephrine is 0 with a patient weight of 62.8 kg, the cardiovascular SOFA is 1.
Given that minimum PO2 is 100.0 and minimum FiO2 is 0.5 the calculated PAO2FIO2 is 200.0, this means the respiratory SOFA is 2.
Because the minimum Platelet count is 310.0 the \hi{coagulation SOFA is 0}.
The maximum Bilirubin (Total) is 1 leading to a \hi{liver SOFA of 0}.
Because total Urine output is 1095.0 and maximum creatinine in the blood is 0.4 the \hi{renal SOFA is 0}.
To summarize: the patient has a total \hi{SOFA score of 5}.
\\~\\
Now we need to calculate the SOFA scores with forecasted values. The SOFA scores in the future based on the forecasted values are the following:
The minimum value of GCS\_eye will be 4.0, GCS\_motor will be 6.0 and GCS\_verbal will be 1.0, this produces the sum 11.0 and means the CNS SOFA will be 2.
Because future minimum MAP will be 65.333, future max Dopamine will be 0, future max Dobutamine will be 0, future max Epinephrine will be 0 and future max Norepinephrine will be 0 with a patient weight of 62.8 kg, the cardiovascular SOFA will be 1.
Given that minimum PO2 will be 100.0 and minimum FiO2 will be 0.5 the forecasted PAO2FIO2 will be 200.0, this means the respiratory SOFA will be 2.
Because the Platelet count will be 310.0 the \hi{coagulation SOFA is going to be 0}.
The maximum Bilirubin (Total) will be 1 leading to a \hi{liver SOFA of 0}.
Because Urine output will be 150.0 and maximum creatinine in the blood will be 0.4 the renal SOFA will be 4.
\hi{To summarize: the patient will have a future total SOFA score of 9.}
\\~\\
\hi{This calculation means that the patient will likely experience a kidney failure since SOFA increased by 4.
The patient will develop sepsis in the next 24 hours, because total SOFA increased by 4 and infection is suspected.}
\end{minipage}
\hfill
\begin{minipage}[t]{0.48\textwidth}
\scriptsize
Patient is 75.0 years old and is male. Given all the information in this text, answer the question at the end.
Here are the measurements: DBP at time -22.37: 49.0, SBP at time -22.37: 105.0, DBP at time -20.37: 52.0, GCS\_eye at time -20.37: 4.0, GCS\_motor at time -20.37: 6.0, GCS\_verbal at time -20.37: 1.0, SBP at time -20.37: 117.0, DBP at time -19.37: 56.0, FiO2 at time -19.37: 0.5, SBP at time -19.37: 127.0, DBP at time -18.37: 43.0,...
\\~\\
Now answer the following question:
\hi{The patient has an existing precondition given by the ICD-9 codes 201.1, 572.8.}
\hi{The patient has ongoing RRT therapy.}

The doctors suspect an infection, based on this information and the other information in this text, will the patient be classified as septic tomorrow?
\vspace{0.5em}
\hrule
\vspace{0.5em}
First we need to calculate the SOFA scores given the extracted values. The SOFA scores for the current time are the following:
The minimum value of GCS\_eye is 4.0, GCS\_motor is 6.0 and GCS\_verbal is 1.0, this produces the sum 11.0 and means the CNS SOFA is 2.
Because minimum MAP is 55.333, max Dopamine is 0, max Dobutamine is 0, max Epinephrine is 0 and max Norepinephrine is 0 with a patient weight of 62.8 kg, the cardiovascular SOFA is 1.
Given that minimum PO2 is 100.0 and minimum FiO2 is 0.5 the calculated PAO2FIO2 is 200.0, this means the respiratory SOFA is 2.
Because the minimum Platelet count is 310.0 the \hi{coagulation SOFA is 4}.
The maximum Bilirubin (Total) is 1 leading to a \hi{liver SOFA of 2}.
Because total Urine output is 1095.0 and maximum creatinine in the blood is 0.4 the \hi{renal SOFA is 4}.
To summarize: the patient has a total \hi{SOFA score of 15}.
\\~\\
Now we need to calculate the SOFA scores with forecasted values. The SOFA scores in the future based on the forecasted values are the following:
The minimum value of GCS\_eye will be 4.0, GCS\_motor will be 6.0 and GCS\_verbal will be 1.0, this produces the sum 11.0 and means the CNS SOFA will be 2.
Because future minimum MAP will be 65.333, future max Dopamine will be 0, future max Dobutamine will be 0, future max Epinephrine will be 0 and future max Norepinephrine will be 0 with a patient weight of 62.8 kg, the cardiovascular SOFA will be 1.
Given that minimum PO2 will be 100.0 and minimum FiO2 will be 0.5 the forecasted PAO2FIO2 will be 200.0, this means the respiratory SOFA will be 2.
Because the Platelet count will be 310.0 the \hi{coagulation SOFA is going to be 4}.
The maximum Bilirubin (Total) will be 1 leading to a \hi{liver SOFA of 2}.
Because Urine output will be 150.0 and maximum creatinine in the blood will be 0.4 the renal SOFA will be 4.
\hi{To summarize: the patient will have a future total SOFA score of 15}.
\\~\\
\hi{The patient will not develop sepsis in the next 24 hours, because total SOFA increased by 0 and infection is  suspected.}
\end{minipage}
    \caption{Finetuning data including a verbalization of Sepsis-3 inference (left column) and inference under an exception due to medical preconditions (right column).  Differences are shown in bold blue font. The general prompt is shown above the horizontal line, the gold standard answer below.} 
    \label{fig:verbalization}
\end{figure*}

Figure \ref{fig:verbalization}, left column, shows an example for finetuning data including inference rules following the Sepsis-3 definition. The data starts with a verbalization of a sparse multivariate input time series of measurements for a 24 hour window, followed by an instruction to classify the patient as septic in the next 24 hours, given information about suspected infections. The third text block in the finetuning data consists of a gold standard inference chain explaining the scores of the six SOFA systems from the corresponding clinical measurements, and computing their sum, given the measurements of the first 24 hour window. The fourth text block includes forecasts of the clinical variables relevant for SOFA for the next 24 hours, and the corresponding inference about the six SOFA subscores, and their  sum. The last text block contains a gold standard inference about the Sepsis prediction and the responsible SOFA system, if applicable. 

Figure \ref{fig:verbalization}, right column, shows an example for finetuning data using exceptions. The format is the same as for inference according to the definition, except for the inclusion of a precondition in the second text block --- in our example, the ICD codes 201.1 (i.e., indicating Hodgkin's granuloma, a lymphoma), 572.8 (i.e., other sequelae of chronic liver disease) and RRT usage --- inducing a change to the SOFA score of the corresponding organ systems.

\newpage

\vspace{0pt}
\subsection{Finetuning Data for SAPS-II}
\label{appendix:finetune-prompt-sapsii}
\noindent

\begin{minipage}[t]{0.98\textwidth}
\scriptsize
Patient is 70.0 years old and is female. Given all the information in this text, answer the question at the end.

Here are the measurements: SBP at time -23.97: 96.0, HR at time -23.1: 84.0, SBP at time -23.1: 93.0, Urine at time -23.1: 80.0, GCS\_eye at time -22.1: 4.0, GCS\_motor at time -22.1: 6.0, GCS\_verbal at time -22.1: 5.0, HR at time -22.1: 79.0, SBP at time -22.1: 97.0, Temperature at time -22.1: 36.278, Urine at time -22.1: 60.0 ...

Now answer the following question:

What is the SAPS-II score of the patient?

\vspace{0.5em}
\hrule
\vspace{0.5em}
We need to calculate the SAPS-II scores given the extracted values.

The patient's age is 70.0, contributing 15 points.

The worst HR value is 79.0, contributing 0 points.

The worst SBP value is 93.0, contributing 5 points.

The worst Temperature value is 36.278, contributing 0 points.

The worst GCS value is 15.0, contributing 0 points.

The worst PaO2/FiO2 ratio is 770.0, contributing 6 points.

The worst BUN value is 15, contributing 0 points.

The total Urine output is 140.0, contributing 11 points.

The worst Sodium value is 138, contributing 0 points.

The worst Potassium value is 4.0, contributing 0 points.

The worst Bicarbonate value is 22, contributing 0 points.

The worst WBC value is 10, contributing 0 points.

The worst Bilirubin value is 1, contributing 0 points.

To summarize: the patient has a total SAPS-II score of 37.

The logit is computed as -7.7631 + 0.0737 * 37 + 0.9971 * ln(37 + 1) = -1.409.

The predicted hospital mortality probability is e\^{}logit / (1 + e\^{}logit) = 0.196.
\end{minipage}

\newpage
\onecolumn
\subsection{One-Shot Prompt for Sepsis-3}
\label{appendix:one-shot}
\begin{minipage}[t]{0.98\textwidth}
\scriptsize
For the given 24 hours of measurements, apply the definitions of the SOFA score given by the following table:

\begin{minted}[breaklines, fontsize=\tiny]{text}
{| class="wikitable"
!
!Central nervous system
!Cardiovascular system
!Respiratory system
!Coagulation
!Liver
!Renal function
|-
!Score
! [[Glasgow coma scale]]
! Mean arterial pressure OR administration of vasopressors required
! PaO<sub>2</sub>/FiO<sub>2</sub> <nowiki>[mmHg (kPa)]</nowiki>
! Platelets (x10<sup>3</sup>/ul)
! Bilirubin (mg/dl) [umol/L]
! Creatinine (mg/dl) [umol/L] (or urine output)
|-
! +0
| 15 || [[Mean_arterial_pressure|MAP]] >= 70 mmHg || >= 400 (53.3) || >= 150 || < 1.2 [< 20] || < 1.2 [< 110]
|-
! +1
| 13-14 || MAP < 70 mmHg || < 400 (53.3) || < 150 || 1.2-1.9 [20-32] || 1.2-1.9 [110-170]
|-
! +2
| 10-12 || [[dopamine]]  <= 5 ug/kg/min or [[dobutamine]] (any dose) || < 300 (40) || < 100 || 2.0-5.9 [33-101] || 2.0-3.4 [171-299]
|-
! +3
| 6-9 || dopamine > 5 ug/kg/min OR [[epinephrine]] <= 0.1 ug/kg/min OR [[norepinephrine]]  <= 0.1 ug/kg/min || < 200 (26.7) '''and''' mechanically ventilated including CPAP || < 50 || 6.0-11.9 [102-204] || 3.5-4.9 [300-440] (or < 500 ml/day)
|-
! +4
| < 6 || dopamine > 15 ug/kg/min OR epinephrine > 0.1 ug/kg/min OR norepinephrine > 0.1 ug/kg/min || < 100 (13.3) '''and''' mechanically ventilated including CPAP || < 20 || > 12.0 [> 204] || > 5.0 [> 440] (or < 200 ml/day)
|}
\end{minted}

A patient is septic if suspected infection is positive and the future total SOFA score calculated with a best guess on how the values develop is two points (or more) higher than the current SOFA score.

Only answer like in the given example. Here is the example:

Patient is 63.0 years old and is male. Given all the information in this text, answer the question at the end.

Here are the measurements: DBP at time -23.68: 36.0, SBP at time -23.68: 71.0...

Now answer the following question in the given format:

Doctors suspect an infection, based on this information and the other information in this text, will the patient be classified as septic tomorrow?

First we need to calculate the SOFA scores given the extracted values. The SOFA scores for the current time are the following:

The minimum value of GCS\_eye is 4.0, GCS\_motor is 6.0 and GCS\_verbal is 5.0, this produces the sum 15.0 and means the CNS SOFA is 0.

Because minimum MAP is 43.333, max Dopamine is 0, max Dobutamine is 0, max Epinephrine is 0 and max Norepinephrine is 0 with a patient weight of 80 kg, the cardiovascular SOFA is 1.

Given that minimum PO2 is 141.0 and minimum FiO2 is 1 the calculated PAO2FIO2 is 141.0, this means the respiratory SOFA is 3.

Because the minimum Platelet count is 235.0 the coagulation SOFA is 0.

The maximum Bilirubin (Total) is 1.8 leading to a liver SOFA of 1.

Because total Urine output is 1585.0 and maximum creatinine in the blood is 1.4 the renal SOFA is 1.

To summarize: the patient has a total SOFA score of 6.

Now we need to calculate the SOFA scores with forecasted values. The SOFA scores in the future based on the forecasted values are the following:

The minimum value of GCS\_eye will be 4.0, GCS\_motor will be 6.0 and GCS\_verbal will be 5.0, this produces the sum 15.0 and means the CNS SOFA will be 0.

Because future minimum MAP will be 55.667, future max Dopamine will be 0, future max Dobutamine will be 0, future max Epinephrine will be 0 and future max Norepinephrine will be 0 with a patient weight of 80 kg, the cardiovascular SOFA will be 1.

Given that minimum PO2 will be 141.0 and minimum FiO2 will be 1 the forecasted PAO2FIO2 will be 141.0, this means the respiratory SOFA will be 3.

Because the Platelet count will be 295.0 the coagulation SOFA is going to be 0.
The maximum Bilirubin (Total) will be 1.8 leading to a liver SOFA of 1.

Because Urine output will be 1635.0 and maximum creatinine in the blood will be 1.1 the renal SOFA will be 0.

To summarize: the patient will have a future total SOFA score of 5.

The patient will not develop sepsis in the next 24 hours, because total SOFA increased only by -1 and infection is suspected.

The example is now finished. Say "The patient will develop sepsis" in the last sentence if the criteria are met (if total SOFA changed by 2 and infection is suspected).

Patient is 76.0 years old and is female. Given all the information in this text, answer the question at the end.

Here are the measurements: DBP at time -22.97: 56.0, GCS\_eye at time -22.97: 4.0, GCS\_motor at time -22.97: 6.0...

Now answer the following question in the given format:

The doctors don't suspect an infection, based on this information and the other information in this text, will the patient be classified as septic tomorrow?
\end{minipage}

\newpage

\subsection{One-Shot Prompt for Sepsis-3 with Exceptions}
\label{appendix:one-shot-exception}
\begin{minipage}[t]{0.98\textwidth}
\scriptsize
For the given 24 hours of measurements, apply the definitions of the SOFA score given by the following table:
For the given 24 hours of measurements, apply the definitions of the SOFA score given by the following table:

\begin{minted}[breaklines, fontsize=\tiny]{text}
{| class="wikitable"
!
!Central nervous system
!Cardiovascular system
!Respiratory system
!Coagulation
!Liver
!Renal function
|-
!Score
! [[Glasgow coma scale]]
! Mean arterial pressure OR administration of vasopressors required
! PaO<sub>2</sub>/FiO<sub>2</sub> <nowiki>[mmHg (kPa)]</nowiki>
! Platelets (x10<sup>3</sup>/ul)
! Bilirubin (mg/dl) [umol/L]
! Creatinine (mg/dl) [umol/L] (or urine output)
|-
! +0
| 15 || [[Mean_arterial_pressure|MAP]] >= 70 mmHg || >= 400 (53.3) || >= 150 || < 1.2 [< 20] || < 1.2 [< 110]
|-
! +1
| 13-14 || MAP < 70 mmHg || < 400 (53.3) || < 150 || 1.2-1.9 [20-32] || 1.2-1.9 [110-170]
|-
! +2
| 10-12 || [[dopamine]]  <= 5 ug/kg/min or [[dobutamine]] (any dose) || < 300 (40) || < 100 || 2.0-5.9 [33-101] || 2.0-3.4 [171-299]
|-
! +3
| 6-9 || dopamine > 5 ug/kg/min OR [[epinephrine]] <= 0.1 ug/kg/min OR [[norepinephrine]]  <= 0.1 ug/kg/min || < 200 (26.7) '''and''' mechanically ventilated including CPAP || < 50 || 6.0-11.9 [102-204] || 3.5-4.9 [300-440] (or < 500 ml/day)
|-
! +4
| < 6 || dopamine > 15 ug/kg/min OR epinephrine > 0.1 ug/kg/min OR norepinephrine > 0.1 ug/kg/min || < 100 (13.3) '''and''' mechanically ventilated including CPAP || < 20 || > 12.0 [> 204] || > 5.0 [> 440] (or < 200 ml/day)
|}
\end{minted}

A patient is septic if suspected infection is positive and the future total SOFA score calculated with a best guess on how the values develop is two points (or more) higher than the current SOFA score.

Only answer like in the given example. Here is the example:

Patient is 74.0 years old and is male. Given all the information in this text, answer the question at the end.
Here are the measurements: Weight at time -23.62: 71.1, DBP at time -23.37: 43.0, GCS\_ eye at time -23.37: 4.0, GCS\_ motor at time -23.37: 6.0...

Now answer the following question:
The patient has an existing precondition given by the ICD-9 codes 403.91, 585.6, 202.0.

The patient has ongoing RRT therapy.

The doctors don't suspect an infection, based on this information and the other information in this text, will the patient be classified as septic tomorrow?

First we need to calculate the SOFA scores given the extracted values. The SOFA scores for the current time are the following:

The minimum value of GCS\_ eye is 4.0, GCS\_ motor is 6.0 and GCS\_ verbal is 5.0, this produces the sum 15.0 and means the CNS SOFA is 0.

Because minimum MAP is 38.667, max Dopamine is 0, max Dobutamine is 0, max Epinephrine is 0 and max Norepinephrine is 0 with a patient weight of 71.1 kg, the cardiovascular SOFA is 1.

Given that minimum PO2 is 770 and minimum FiO2 is 1 the calculated PAO2FIO2 is 770.0, this means the respiratory SOFA is 0.

Because the minimum Platelet count is 49.0 the coagulation SOFA is 4.
The maximum Bilirubin (Total) is 0.2 leading to a liver SOFA of 0.

Because total Urine output is 0 and maximum creatinine in the blood is 3.9 the renal SOFA is 4.

To summarize: the patient has a total SOFA score of 9.

Now we need to calculate the SOFA scores with forecasted values. The SOFA scores in the future based on the forecasted values are the following:

The minimum value of GCS\_ eye will be 4.0, GCS\_ motor will be 6.0 and GCS\_ verbal will be 5.0, this produces the sum 15.0 and means the CNS SOFA will be 0.

Because future minimum MAP will be 37.0, future max Dopamine will be 0, future max Dobutamine will be 0, future max Epinephrine will be 0 and future max Norepinephrine will be 0 with a patient weight of 71.1 kg, the cardiovascular SOFA will be 1.

Given that minimum PO2 will be 770 and minimum FiO2 will be 1 the forecasted PAO2FIO2 will be 770.0, this means the respiratory SOFA will be 0.

Because the minimum Platelet count will be 37.0 the coagulation SOFA is going to be 4.

The maximum Bilirubin (Total) will be 0.3 leading to a liver SOFA of 0.

Because Urine output will be 0 and maximum creatinine in the blood will be 5.3 the renal SOFA will be 4.

To summarize: the patient will have a future total SOFA score of 9.

The patient will not develop sepsis in the next 24 hours, because total SOFA changed only by 0 and infection is not suspected

The example is now finished. Say "The patient will develop sepsis" in the last sentence if the criteria are met (if total SOFA changed by 2 and infection is suspected).

Here are the measurements: DBP at time -22.97: 56.0, GCS\_eye at time -22.97: 4.0, GCS\_motor at time -22.97: 6.0...

The patient has an existing precondition given by the ICD-9 codes 206.10, 572.3.

Now answer the following question in the given format:

The doctors don't suspect an infection, based on this information and the other information in this text, will the patient be classified as septic tomorrow?
\end{minipage}

\newpage

\subsection{One-Shot Prompt for SAPS-II}
\label{appendix:one-shot-sapsii}

\begin{minipage}[h]{0.98\textwidth}
\scriptsize
For the given 24 hours of measurements, apply the definitions of the SAPS-II score given by the following table:

\begin{minted}[breaklines, fontsize=\tiny]{text}

\begin{table}[H]
        \begin{threeparttable}[b]
        \centering
        \small
        \caption{List of  non-static components of SAPS-II score \citep{LeGallETAL:93}. 
                        The total score is computed by summing the individual contributions. 
                        Abbreviations: HR = heart rate; SBP = systolic blood pressure;  GCS = Glasgow Coma Scale; BUN = blood urea nitrogen; WBC = white blood cell count, MV = mechanically ventilated including CPAP.}
        \begin{tabular}{lcccccccccccc}
                \toprule
                Score & HR\tnote{1} & SBP\tnote{1} &  Temp\tnote{2}  & GCS\tnote{3} & PaO2/Fi02\tnote{3} & BUN\tnote{2} & Urine\tnote{4} & Sodium\tnote{1} & Potassium\tnote{1} & Bicarbonate\tnote{3} & WBC\tnote{1} & Bilirubin\tnote{2} \\
                      &    [1/s]    &    [mmHG]    & [\textdegree C] &              &     [mmHg/\%]      &   [mg/dl]    &    [ml/24h]    &    [mmol/L]     &      [mEq/L]       &       [mEq/L]        &  [1000/mm³]  &      [mg/dL]       \\ \midrule
                +0    &   70-119    &   100-199    &      else       &    14-15     &       not MV       &     $<$28      &   $\geq$1000   &     125-144     &      3.0-4.9       &       $\geq$20       &   1.0-19.9   &       $<$4.0       \\
                +1    &             &              &                 &              &                    &              &                &    $\geq$145    &                    &                      &              &                    \\
                +2    &    40-69    &    $>$199    &                 &              &                    &              &                &                 &                    &                      &              &                    \\
                +3    &             &              &    $\geq$39     &              &                    &              &                &                 &                    &        15-19         &   $\geq$20   &                    \\
                +4    &   120-159   &              &                 &              &                    &              &    500-999     &                 &                    &                      &              &      4.0-5.9       \\
                +5    &             &    70-99     &                 &    11-13     &                    &              &                &     $<$125      &  $<$3 or $\geq$5   &                      &              &                    \\
                +6    &             &              &                 &              &       $>$199       &    28-83     &                &                 &                    &        $<$15         &              &                    \\
                +7    &   $>$159    &              &                 &     9-10     &                    &              &                &                 &                    &                      &              &                    \\
                +9    &             &              &                 &              &      100-199       &              &                &                 &                    &                      &              &      $\geq$6       \\
                +10   &             &              &                 &              &                    &    $>$83     &                &                 &                    &                      &              &                    \\
                +11   &    $<$40    &              &                 &              &       $<$100       &              &     $<$500     &                 &                    &                      &              &                    \\
                +12   &             &              &                 &              &                    &              &                &                 &                    &                      &    $<$1.0    &                    \\
                +13   &             &    $<$70     &                 &     6-8      &                    &              &                &                 &                    &                      &              &                    \\
                +26   &             &              &                 &     $<$6     &                    &              &                &                 &                    &                      &              &                    \\ \bottomrule
                      &             &              &                 &              &
        \end{tabular}
        \begin{tablenotes}
                \item [1] worst (min or max) value in 24h
                \item [2] largest (max) value in 24h
                \item [3] smallest (min) value in 24h
                \item [4] total (sum) value in 24h
        \end{tablenotes}        
\end{threeparttable}
\end{table}
\end{minted}

For age, use the following thresholds: <40: 0, 40-59: 12, 60-69: 15, 70-74: 16, >= 80: 18 

Only answer like in the given example. Here is the example:

Patient is 70.0 years old and is female. Given all the information in this text, answer the question at the end.

Here are the measurements: SBP at time -23.97: 96.0, HR at time -23.1: 84.0...
Now answer the following question:
What is the SAPS-II score of the patient?
We need to calculate the SAPS-II scores given the extracted values.
The patient's age is 70.0, contributing 15 points.
The worst HR value is 79.0, contributing 0 points.
The worst SBP value is 93.0, contributing 5 points.
The worst Temperature value is 36.278, contributing 0 points.
The worst GCS value is 15.0, contributing 0 points.
The worst PaO2/FiO2 ratio is 770.0, contributing 6 points.
The worst BUN value is 15, contributing 0 points.
The total Urine output is 140.0, contributing 11 points.
The worst Sodium value is 138, contributing 0 points.
The worst Potassium value is 4.0, contributing 0 points.
The worst Bicarbonate value is 22, contributing 0 points.
The worst WBC value is 10, contributing 0 points.
The worst Bilirubin value is 1, contributing 0 points.
To summarize: the patient has a total SAPS-II score of 37.
The logit is computed as -7.7631 + 0.0737 * 37 + 0.9971 * ln(37 + 1) = -1.409.
The predicted hospital mortality probability is e\^{}logit / (1 + e\^{}logit) = 0.196.

The example is now finished.

Here are the measurements: HR at time -23.1: 79.0, SBP at time -23.1: 97.0, Temperature at time -23.1: 36.278 ...

Now answer the following question in the given format:
What is the SAPS-II score of the patient?

\end{minipage}

\newpage

\subsection{Additional Analyses}
\label{appendix:further-analyses}

\paragraph{Reproducing the non-linear SAPS-II logit.} The SAPS-II step is the one place where the consensus rule leaves the step-function regime for a logistic transform (Section~\ref{sec:inference}). The finetuned model reproduces this logit-and-sigmoid mapping with derivation correctness $\approx 1.0$ (Table~\ref{tab:results_inference}) and value correctness $\approx 0.99$ on both logit and probability (Table~\ref{tab:results_value_sofa}), whereas every one-shot baseline collapses on this step (e.g.\ Total SAPS-II value correctness $\leq 0.14$) \emph{even though the exact formula is supplied in the prompt}. Reproducing a logistic transform from worked examples rather than from a stated equation is exactly the compositional, non-linear inference that finetuning unlocks and one-shot prompting does not.

\paragraph{Guideline-free direct predictors reach a similar F1 range.} To test whether the modest Sepsis-3 F1 is a property of the 24-hour prognostic horizon rather than of the guideline scaffolding, we trained a model that bypasses the guideline chain entirely and predicts the Sepsis-3 label directly from the same 24-hour input window. Its architecture and meta-parameters are those of the multimodal model (Appendices~\ref{appendix:tsf} and~\ref{appendix:hyperparameters}). The only differences are a classification head instead of the text-based decoder, and the Sepsis-3 label instead of the inference chain as the supervision signal. This also serves as an ablation of inference-chain supervision against answer-only supervision. On the same test split it obtains F1 $0.277$ (AUROC $0.820$), in the same neighborhood as both the text-only finetuned ($0.254$) and multimodal ($0.309$) guideline models.

For another baseline, we trained XGBoost on the same task, using the default settings except for 1500 estimators, depth 6, \texttt{min\_child\_weight} 5, learning rate 0.01, \texttt{scale\_pos\_weight} set to the ratio of negative to positive examples, early stopping after 150 rounds, and \texttt{colsample\_bytree} 0.5. On the full 24-hour input matrix it obtains F1 $0.300$ (AUROC $0.831$). With engineered temporal features (mean, minimum, maximum, and count in the 0--6h, 6--12h, and 12--24h windows, and deltas between the means) it obtains F1 $0.332$ (AUROC $0.856$). All approaches stay in a similar F1 range, which supports the claim that the forecasting horizon limits performance more than the  architecture does.

These AUROCs are comparable to direct sepsis predictors in the literature, which report AUC around $0.83$ at substantially shorter (e.g.\ 12-hour) horizons on MIMIC-III, often on balanced sets \citep{NematiETAL:18}; our horizon is 24 hours.

\newpage

\subsection{Comparison with Further One-Shot LLMs}
\label{appendix:further-llms}

To address whether our findings extend to recent medical and reasoning-tuned models, we ran \texttt{FreedomIntelligence/HuatuoGPT-o1-8B} and \texttt{allenai/Olmo-3-7B-Instruct} with the exact one-shot prompts of Appendices~\ref{appendix:one-shot}--\ref{appendix:one-shot-sapsii}. Neither adheres to the prescribed template, in contrast to all one-shot baselines reported in the main paper.

\paragraph{Olmo-3-7B-Instruct} produces token-level incoherent output, so parsing fails and our evaluation metrics cannot be applied. A representative fragment:
\begin{quote}\small\itshape
, the minimum 6.0. If the future GCS\_eye: 4, GCS\_motor at.0, GCS\_verbal: 5.0, the maximum 7.0. If GCS at -7.0, then the future SBP at -7.83: 150.0, Urine at -7.83: 45.0, the minimum. If the patient is 75, the future.0, and the future eye is 4.0, then the patient will be septic.
\end{quote}

\paragraph{HuatuoGPT-o1-8B} emits fluent reasoning in its own RL-induced ``Thinking / Final Response'' format, but the trace is internally incoherent and contradicts the consensus rule. In the example below (a patient who did \emph{not} develop sepsis), the model (i)~calls the difference between $4$ and $6$ a one-point increase when it is clearly two, and (ii)~contradicts the Sepsis-3 criterion it had just stated, claiming that suspicion of infection alone overrides the SOFA-change requirement, which it does not. This is exactly the failure mode our process-level evaluation is designed to surface: fluent but unfaithful reasoning. Because the model reliably ends with ``the patient will (not) develop sepsis,'' we can score Sepsis-3 final-answer accuracy only; it obtains accuracy $0.765$ and F1 $0.204$, below the one-shot LLaMA-3-8B baseline (accuracy $0.834$, F1 $0.231$) despite its medical-RL pretraining and substantially longer traces. Derivation and value correctness on intermediate nodes are undefined because the trace does not follow the prescribed structure.

\begin{quote}\small\itshape
\textbf{Thinking.} Alright, let's figure out what's happening here. \dots First, I need to look at their current SOFA score. Their CNS score is 0 \dots cardiovascular \dots a +0. \dots respiration \dots another +0. Coagulation \dots +0 \dots liver \dots renal \dots alright. So, adding everything up, their current SOFA score is 4. Next \dots If we assume the worst, their future SOFA score could go up to 6. That's a one-point increase from today \dots not enough \dots since we'd need a two-point increase. But wait \dots if doctors suspect an infection \dots even with just a one-point increase in SOFA, they could still be considered septic. \dots the patient would likely be classified as septic tomorrow.\\[2pt]
\textbf{Final Response.} The patient will develop sepsis. Based on the provided data, the patient's current SOFA score is 4. Assuming the worst-case scenario for tomorrow, the potential future SOFA score could reach 6. Although this represents a one-point increase rather than the necessary two-point increase for sepsis classification, the suspicion of an infection overrides this requirement. Therefore, the patient will be classified as septic tomorrow.
\end{quote}

\begin{table*}[h!]
    \centering
    \begin{tabular}{lll}
    \toprule
    Model & Sepsis-3 Accuracy & Sepsis-3 F1 \\
    \midrule
    one-shot LLaMA-3-8B (baseline) & 0.834 & 0.231 \\
    HuatuoGPT-o1-8B                & 0.765 & 0.204 \\
    Olmo-3-7B-Instruct             & \multicolumn{2}{c}{output incoherent; metrics undefined} \\
    \bottomrule
    \end{tabular}
    \caption{Final-answer Sepsis-3 performance of additional recent one-shot LLMs. Intermediate derivation/value correctness cannot be computed because these models do not follow the prescribed inference template.}
    \label{tab:further_llms}
\end{table*}

\newpage

\subsection{Time Series Forecasting}
\label{appendix:tsf}

Our TSF model uses the implementation of \cite{StaniekETAL:24mlhc} which is based on a Transformer encoder-decoder architecture \citep{VaswaniETAL:17}. First, sparse multivariate input time series are represented as quadruplets $\text{S} = \{(f_i, t_i, v_i, n_i)\}_{i=1}^n$, where $f_i \in F$ is a clinical variable identifier, $t_i \in \mathbb{R}_{\geq 0}$ is a time index, $v_i \in \mathbb{R}$ the observed value of $f_i$ at $t_i$, and $n_i$ the unique stay identifier. Then the quadruplets for a 24 hour time series are encoded into a dense representation $x$ where every timestep is a vector of feature values representing one hour.
We construct this vector by choosing the first observed value during the represented hour for each feature. If no value was observed, we impute zero which corresponds to the mean value due to standardization of the data.
Additionally, a mask indicating whether a value was imputed is generated and appended to the vector.
For TSF, we use a Transformer model with an autoregressive iterative multistep (IMS) decoder that generates an output vector $\hat{y}_t \in \mathbb{R}^{|F|}$.
The predicted output $\hat{y}_t$ is a function of the history $\hat{y}_{<t}$ of predicted timesteps until time $t$, the encoded input $x$, and the model parameters $\theta$:
$
\label{eq:autoreg}
    \hat{y}_t = f_\theta(\hat{y}_{<t}, x)
$.
To perform long-term TSF using the autoregressive setup, the outputs $\hat{y}_t$ from each time step $t = 1, \ldots, T$ are concatenated. The complete model is trained with masked mean squared error (MSE).

\paragraph{Multimodal coupling.} For the multimodal variant, the full decoder prediction of the forecaster is extracted and each output vector is passed through a connector MLP that maps it to the Llama-3 embedding dimensionality. The resulting time-series ``token embeddings'' are prepended to the text embeddings of the verbalized prompt. During finetuning, the LORA adapters of the Llama-3 model, the connector MLP, and the TSF model are updated jointly, whereas in the pipeline variant the forecaster is kept fixed and only its predictions are inserted into the prompt.

\begin{figure*}[h!]
    \centering
    \includegraphics[width=0.55\textwidth]{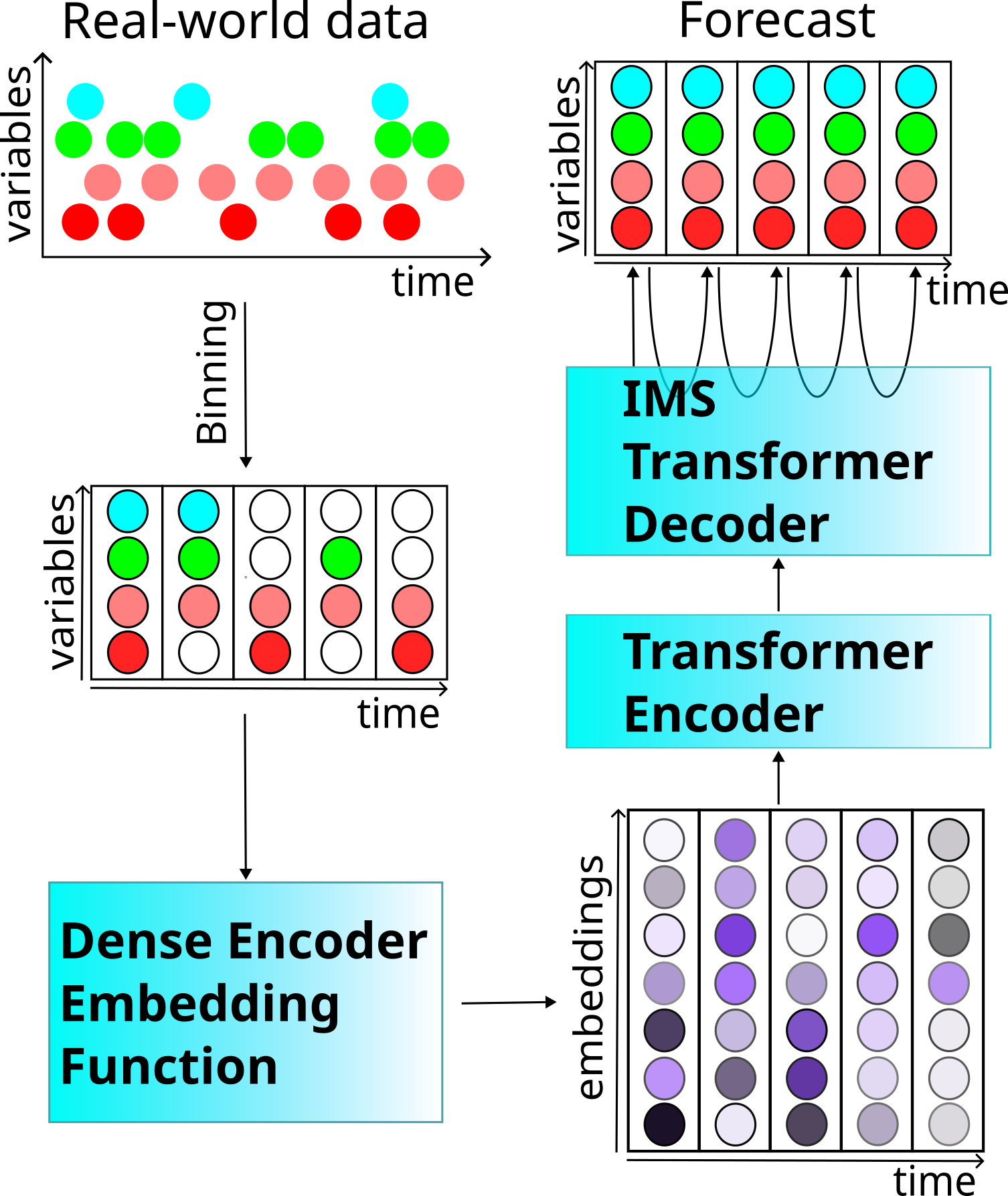}
    \caption{Time Series Forecasting using a dense encoder and iterative multistep decoder architecture.} 
    \label{fig:architecture}
\end{figure*}

\newpage 

\subsection{Hyperparameters and Compute Infrastructure}
\label{appendix:hyperparameters}

\begin{table*}[h!]
    \centering
    \begin{tabular}{ll}
    \toprule
Hyperparameter & value \\
\midrule
Embedding Size & 512 \\
Hidden size encoder & 512 \\
Hidden size decoder & 512 \\
\# Encoder layers & 2 \\
\# Decoder layers & 1 \\
learning rate & 0.0005\\
attention heads encoder & 8\\
attention heads decoder & 1\\
dropout & 0.05\\
max epochs & 100\\
patience (early stopping) & 6 \\
Random Seed & unixtime variation \\
\# GPUs & 1\\
Training all clock time & 5 hours\\
GPU type & Nvidia GTX 1080 Ti\\
\bottomrule
    \end{tabular}
     \caption{Hyperparameter settings of the time series forecaster.}
    \label{tab:hyperparameters}
\end{table*}

\begin{table*}[h]
    \centering
    \begin{tabular}{ll}
    \toprule
Hyperparameter & value \\
\midrule
optimizer & adam \\
learning rate & 0.000002 \\
max epochs & 10  \\
connector first layer & 131x2096\\
connector second layer & 2096x4096\\
lora-r & 16 \\
lora-alpha & 16 \\
lora-dropout &  0.1 \\
lora-target-modules & all-linear \\
\# GPUs & 1\\
Training all clock time & 60 hours\\
GPU type & NVIDIA A100\\
Starting model & meta-llama/Llama-3.1-8B-Instruct\\
\bottomrule
    \end{tabular}
     \caption{Hyperparameter settings of the multi-modal and LLM finetuning architectures.}
\end{table*}

\paragraph{Computational cost.} One training run of our model took 60 GPU hours on one NVIDIA A100 GPU. At 0.4\,kW per GPU this amounts to 24\,kWh, or 30\,kWh including CPU and 35\,kWh including cooling. Assuming a grid carbon intensity of 400\,g CO$_2$e/kWh, this corresponds to 14\,kg CO$_2$e per training run.

\newpage

\subsection{Discussion of 5\% Error Margin}
\label{appendix:margin}

\begin{figure*}[h!]
    \centering
    \includegraphics[width=0.85\linewidth]{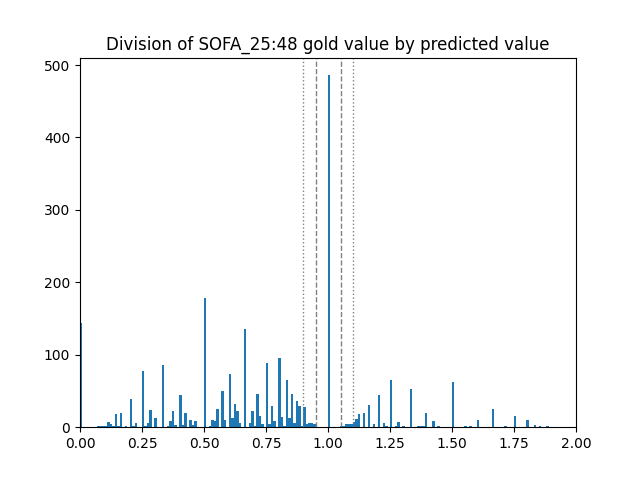}
    \caption{Histogram of the scores obtained by dividing predicted by gold standard values for $\text{SOFA}_{25:48}$ score. Dashed vertical lines show the 5\% interval, dotted vertical lines show the 10\% interval}
    \label{fig:histogram_total_SOFA}
\end{figure*}

The 5\% margin for deviations of predictions is justified by the statistics from our experimental results. For example, $\text{SOFA}_{25:48}$ score ranges between 0 and 24. The experimental results in Figure \ref{fig:histogram_total_SOFA} plot scores obtained by dividing predicted values by gold standard values for this metric, showing that a 10\% margin (dotted vertical lines) would include false positives, while a 5\% margin does not. Similar results are obtained for other predictions.

\newpage

\subsection{Further Evaluation Results for Data with Exceptions to SOFA Rules}
\label{appendix:results_exc}

\begin{table*}[h!]
    \centering
    \resizebox{0.75\textwidth}{!}{
    \begin{tabular}{lllllll|l}
    \toprule
 & variable & one-shot & one-shot-70B & me-llama & deepseek & qwen3.6-27B & finetuned \\
\midrule
\multirow{16}{*}{\begin{sideways}current\end{sideways}} 
& GCS-eye        & 0.849 & 0.981 & 0.932 & 0.807 & 0.985 & 0.998 \\
& GCS-motor      & 0.913 & 0.976 & 0.936 & 0.829 & 0.984 & 0.998 \\
& GCS-verbal     & 0.950 & 0.984 & 0.939 & 0.874 & 0.984 & 0.999 \\
& MAP            & 0.031 & 0.143 & 0.021 & 0.055 & 0.310 & 0.994 \\
& Dopamine       & 0.950 & 0.951 & 0.949 & 0.062 & 0.953 & 0.972 \\
& Dobutamine     & 0.981 & 0.978 & 0.985 & 0.065 & 0.979 & 0.993 \\
& Epinephrine    & 0.968 & 0.965 & 0.973 & 0.065 & 0.966 & 0.979 \\
& Norepinephrine & 0.877 & 0.875 & 0.876 & 0.062 & 0.880 & 0.909 \\
& Weight         & 0.601 & 0.570 & 0.574 & 0.429 & 0.582 & 0.999 \\
& PaO2/FiO2      & 0.494 & 0.749 & 0.393 & 0.340 & 0.807 & 0.998 \\
& PaO2           & 0.829 & 0.769 & 0.814 & 0.669 & 0.833 & 0.998 \\
& FiO2           & 0.887 & 0.811 & 0.897 & 0.778 & 0.894 & 1.000 \\
& Platelet       & 0.917 & 0.956 & 0.840 & 0.686 & 0.968 & 0.997 \\
& Bilirubin      & 0.290 & 0.289 & 0.290 & 0.234 & 0.288 & 0.999 \\
& Urine          & 0.065 & 0.139 & 0.083 & 0.076 & 0.280 & 0.972 \\
& Creatinine     & 0.860 & 0.963 & 0.836 & 0.659 & 0.959 & 0.998 \\
\midrule
\multirow{16}{*}{\begin{sideways}future\end{sideways}} 
& GCS-eye        & 0.673 & 0.708 & 0.676 & 0.548 & 0.634 & 0.705 \\
& GCS-motor      & 0.779 & 0.804 & 0.791 & 0.640 & 0.736 & 0.810 \\
& GCS-verbal     & 0.812 & 0.836 & 0.827 & 0.676 & 0.776 & 0.844 \\
& MAP            & 0.039 & 0.090 & 0.085 & 0.061 & 0.202 & 0.295 \\
& Dopamine       & 0.968 & 0.963 & 0.957 & 0.731 & 0.949 & 0.978 \\
& Dobutamine     & 0.986 & 0.981 & 0.989 & 0.712 & 0.970 & 0.993 \\
& Epinephrine    & 0.979 & 0.971 & 0.977 & 0.701 & 0.956 & 0.989 \\
& Norepinephrine & 0.891 & 0.885 & 0.896 & 0.660 & 0.859 & 0.914 \\
& Weight         & 0.587 & 0.560 & 0.564 & 0.422 & 0.564 & 0.913 \\
& PaO2/FiO2      & 0.328 & 0.328 & 0.320 & 0.286 & 0.419 & 0.563 \\
& PaO2           & 0.462 & 0.365 & 0.502 & 0.416 & 0.457 & 0.617 \\
& FiO2           & 0.747 & 0.635 & 0.751 & 0.646 & 0.730 & 0.825 \\
& Platelet       & 0.261 & 0.323 & 0.311 & 0.222 & 0.323 & 0.363 \\
& Bilirubin      & 0.147 & 0.156 & 0.163 & 0.156 & 0.125 & 0.812 \\
& Urine          & 0.091 & 0.080 & 0.088 & 0.081 & 0.094 & 0.129 \\
& Creatinine     & 0.325 & 0.313 & 0.388 & 0.252 & 0.327 & 0.404 \\
\bottomrule
    \end{tabular}
}
\caption{Value correctness of predicted clinical variables compared to real-world measurements with data-driven exceptions to SOFA rules.}
    \label{tab:results_value_clinical_exception}
\end{table*}

\begin{table*}[h!]
    \centering
\resizebox{0.75\textwidth}{!}{
    \begin{tabular}{lllllll|l}
    \toprule
 & variable & one-shot & one-shot-70B & me-llama & deepseek & qwen3.6-27B & finetuned \\
\midrule
\multirow{7}{*}{\begin{sideways}current\end{sideways}} 
& S$_\text{CNS}$     & 0.727 & 0.773 & 0.603 & 0.385 & 0.974 & 0.998 \\
& S$_\text{cardio}$  & 0.380 & 0.531 & 0.311 & 0.306 & 0.653 & 0.984 \\
& S$_\text{resp}$    & 0.605 & 0.377 & 0.505 & 0.395 & 0.700 & 0.995 \\
& S$_\text{coag}$    & 0.573 & 0.658 & 0.554 & 0.280 & 0.799 & 0.999 \\
& S$_\text{liver}$   & 0.625 & 0.843 & 0.858 & 0.573 & 0.923 & 1.000 \\
& S$_\text{renal}$   & 0.463 & 0.651 & 0.577 & 0.274 & 0.827 & 0.998 \\
& $\text{SOFA}_{1:24}$ & 0.152 & 0.145 & 0.120 & 0.052 & 0.407 & 0.979 \\
\midrule
\multirow{8}{*}{\begin{sideways}future\end{sideways}} 
& S$_\text{CNS}$     & 0.583 & 0.597 & 0.546 & 0.351 & 0.609 & 0.703 \\
& S$_\text{cardio}$  & 0.401 & 0.491 & 0.336 & 0.251 & 0.581 & 0.721 \\
& S$_\text{resp}$    & 0.455 & 0.340 & 0.440 & 0.313 & 0.611 & 0.746 \\
& S$_\text{coag}$    & 0.471 & 0.652 & 0.531 & 0.189 & 0.674 & 0.855 \\
& S$_\text{liver}$   & 0.627 & 0.817 & 0.838 & 0.564 & 0.866 & 0.953 \\
& S$_\text{renal}$   & 0.371 & 0.569 & 0.502 & 0.241 & 0.619 & 0.538 \\
& $\text{SOFA}_{25:48}$ & 0.107 & 0.116 & 0.100 & 0.059 & 0.193 & 0.244 \\
& $\text{SOFA}_\text{diff}$ & 0.089 & 0.093 & 0.083 & 0.082 & 0.235 & 0.107 \\
\midrule
 & SEPSIS Accuracy    & 0.553 & 0.918 & 0.758 & 0.900 & 0.887 & 0.889 \\
 & SEPSIS Specificity & 0.658 & 0.987 & 0.788 & 0.957 & 0.952 & 0.931 \\
 & SEPSIS Sensitivity & 0.136 & 0.019 & 0.396 & 0.105 & 0.142 & 0.349 \\
 & SEPSIS F1          & 0.129 & 0.030 & 0.200 & 0.123 & 0.160 & 0.311 \\
\bottomrule
    \end{tabular}
}
    \caption{Value correctness of predicted SOFA and SEPSIS scores compared to derivations from real-world clinical measurements with data-driven exceptions to SOFA rules.} 
    \label{tab:results_value_sofa_exception}
\end{table*}

\end{document}